\documentclass[10pt,twocolumn,letterpaper]{article}

\usepackage[pagenumbers]{cvpr} 

\usepackage{graphicx}
\usepackage{amsmath}
\usepackage{amssymb}
\usepackage{booktabs}
\usepackage{multirow}
\usepackage{makecell}
\usepackage{amsthm,amsmath,amssymb}
\usepackage{mathrsfs}
\usepackage[pagebackref,breaklinks,colorlinks]{hyperref}
\usepackage{float}
\usepackage{graphicx}
\usepackage{epstopdf}
\usepackage{caption}
\usepackage{subcaption, enumitem}

\usepackage[capitalize]{cleveref}
\crefname{section}{Sec.}{Secs.}
\Crefname{section}{Section}{Sections}
\Crefname{table}{Table}{Tables}
\crefname{table}{Tab.}{Tabs.}

\usepackage{setspace}
\def\cvprPaperID{5945} 
\def\confName{CVPR}
\def\confYear{2023}

\title{DNeRV: Modeling Inherent Dynamics via Difference Neural Representation for Videos}

\author{Qi Zhao\\
Nanjing University\\
{\tt\small qizhao@smail.nju.edu.cn}
\and
M. Salman Asif\\
University of California Riverside\\
{\tt\small sasif@ucr.edu}
\and
Zhan Ma\footnote[1]\\ \\
Nanjing University\\
{\tt\small mazhan@nju.edu.cn}
}

\begin{document}
\twocolumn[{%
\renewcommand\twocolumn[1][]{#1}%
\maketitle
\begin{center}
  \centering
   \includegraphics[width=0.95\linewidth]{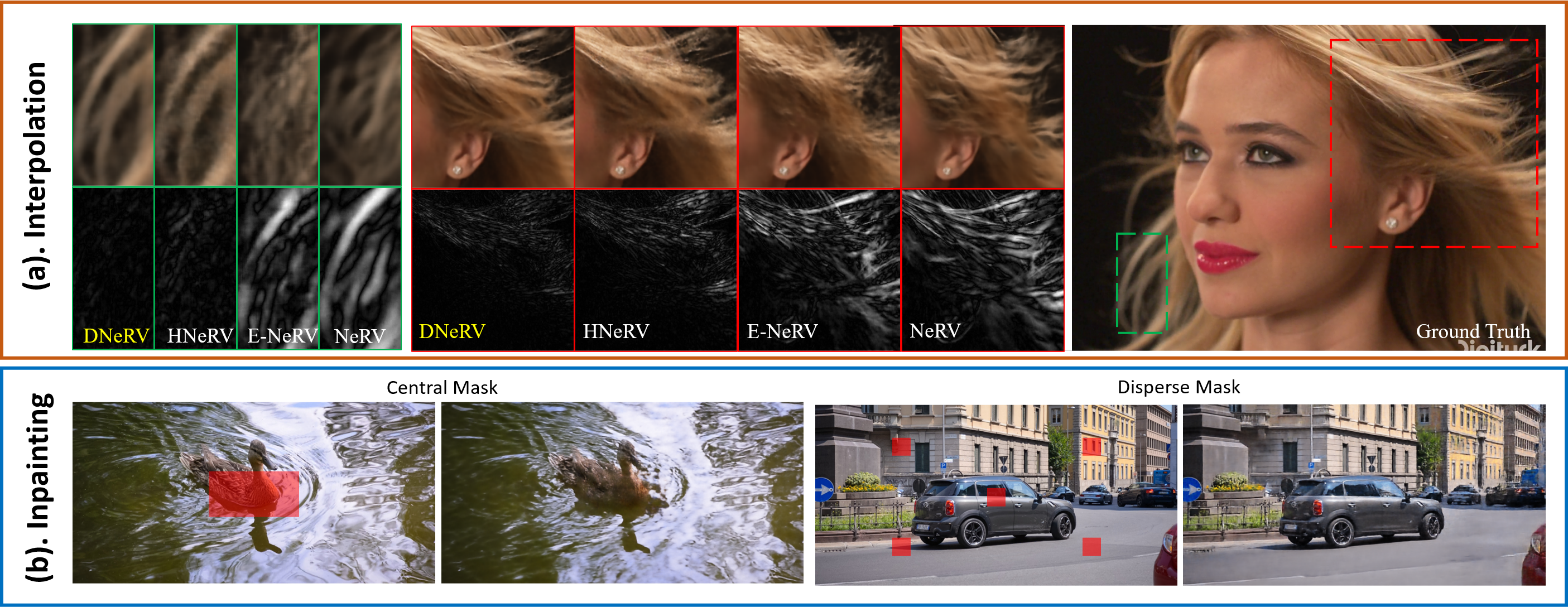}
    \setlength{\belowcaptionskip}{3pt}
   \captionof{figure}{Results of the proposed DNeRV with 3M parameters for (a) video interpolation on UVG~\cite{uvg} and (b) video inpainting on Davis~\cite{davis}. The superior performance shows the effectiveness and generalization capability of DNeRV on neural representation for videos.} 
   \label{fig1}
\end{center}%
}]

\footnotetext[1]{Corresponding author: Zhan Ma (\tt\small mazhan@nju.edu.cn)}
\begin{abstract}
Existing implicit neural representation (INR) methods do not fully exploit spatiotemporal redundancies in videos. Index-based INRs ignore the content-specific spatial features and hybrid INRs ignore the contextual dependency on adjacent frames, leading to poor modeling capability for scenes with large motion or dynamics. We analyze this limitation from the perspective of function fitting and reveal the importance of frame difference. To use explicit motion information, we propose Difference Neural Representation for Videos (DNeRV), which consists of two streams for content and frame difference. We also introduce a collaborative content unit for effective feature fusion. We test DNeRV for video compression, inpainting, and interpolation. DNeRV achieves competitive results against the state-of-the-art neural compression approaches and outperforms existing implicit methods on downstream inpainting and interpolation for $960 \times 1920$ videos.
\end{abstract}

\section{Introduction}
\label{sec:intro}

In recent years, implicit neural representations (INR) have gained significant attention due to their strong ability in  learning a coordinate-wise mapping of different functions. The main principle behind INR is to learn an implicit continuous mapping $f$ using a learnable neural network $g_{\theta}(\cdot):\mathbb{R}^m \rightarrow \mathbb{R}^n$. The idea was first proposed for the neural radiance fields (NeRF)~\cite{nerf} and since then has been applied to various applications \cite{DBLP:conf/eccv/Fan0WGXW22, DBLP:conf/iccv/ZhiLLD21, nerv}.

INR attempts to approximate the continuous $f$ by training $g_{\theta}$ with $m$-dimensional discrete \textit{coordinates} $\mathbf{x} \in \mathbb{R}^m$ and corresponding quantity of interest $\mathbf{y} \in \mathbb{R}^n$. Once trained, the desired $f$ can be fully characterized using $g_\theta$ or the weights $\theta$, and it would be benefit for the tasks which need to model the intrinsic generalization for given data, such as interpolation or inpainting tasks shown in Fig.~\ref{fig1}.

\begin{figure}[t]
  \centering
   \includegraphics[width=0.9\linewidth]{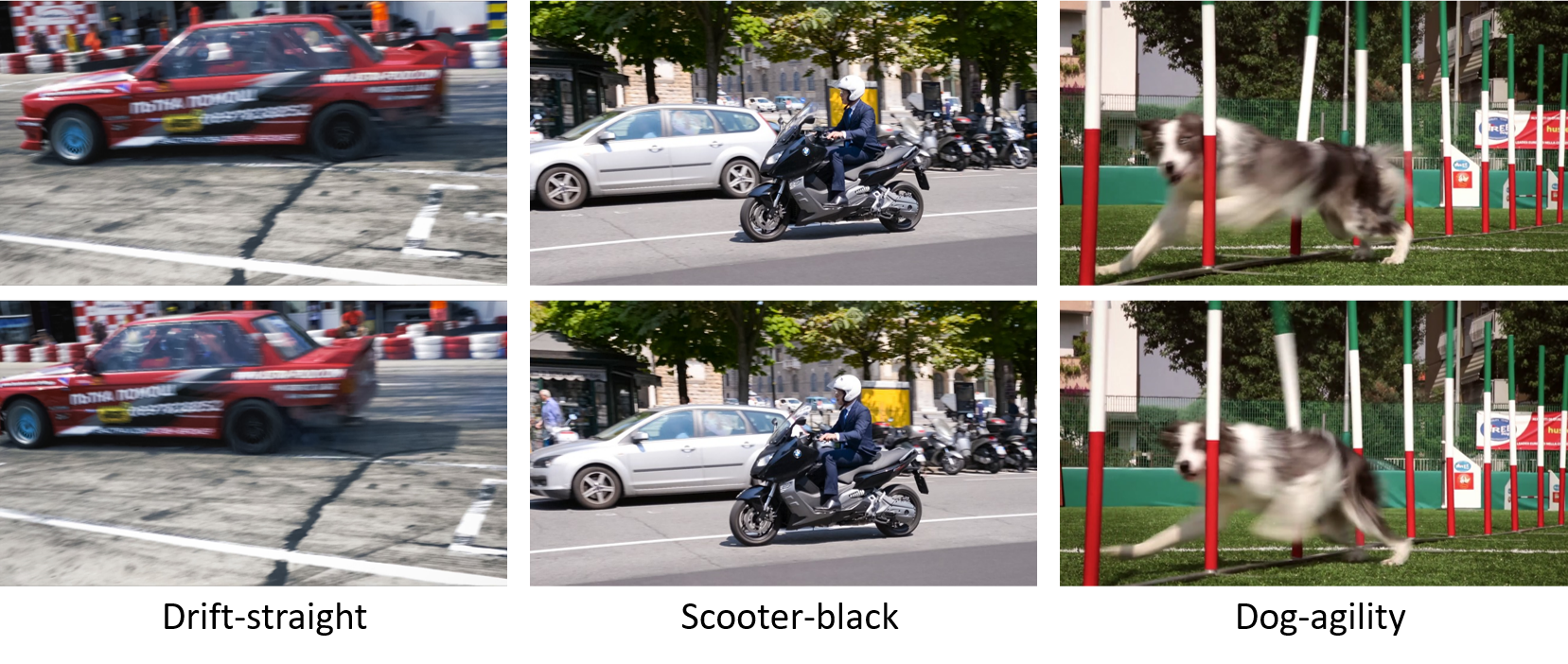}
   \captionof{figure}{Examples of neighboring frames with large mismatch. Learning continuous INR with such dynamics is challenging.}
   \label{davis_sample}
\end{figure}

The success of INR can be attributed to the insight that a learnable and powerful operator with a finite set of data samples $\mathcal{S}=\{ {x_i,y_i} \}^{N}_{i=0}$, can fit the \textit{unknown} mapping $f$. The accuracy of the mapping depends on the number of samples $(N)$ and the complexity of the map $f$. INR for videos requires a large $N$, which primarily depends on the size and internal complexity of the video sequence.
Furthermore, video representation is complicated due to different sampling or frames-per-second (FPS) rates of videos. Large motion (in terms of direction, speed, rotation, or blur) and transformations of the objects or scene can make adjacent frames quite different. Figure~\ref{davis_sample} shows examples of such mismatch between consecutive frames, which we attribute to \textit{adjacent dynamics}.

Adjacent dynamics are the short-term transformations in the spatial structure, which are difficult to represent using existing methods for neural representation of videos (NeRV). Existing NeRV approaches can be broadly divided into two groups: (1) \textit{Index-based} methods, such as \cite{nerv} and \cite{e-nerv}, use positional embedding of the index as input and lack content-specific information for given videos. (2) \textit{Hybrid-based} methods~\cite{HNeRV22} use frames for index embedding and neglect the temporal correlation between different frames. Therefore, neither index nor frame-based NeRV are effective against adjacent dynamics.

\begin{figure*}[t]
  \centering
  \includegraphics[width=0.8\linewidth]{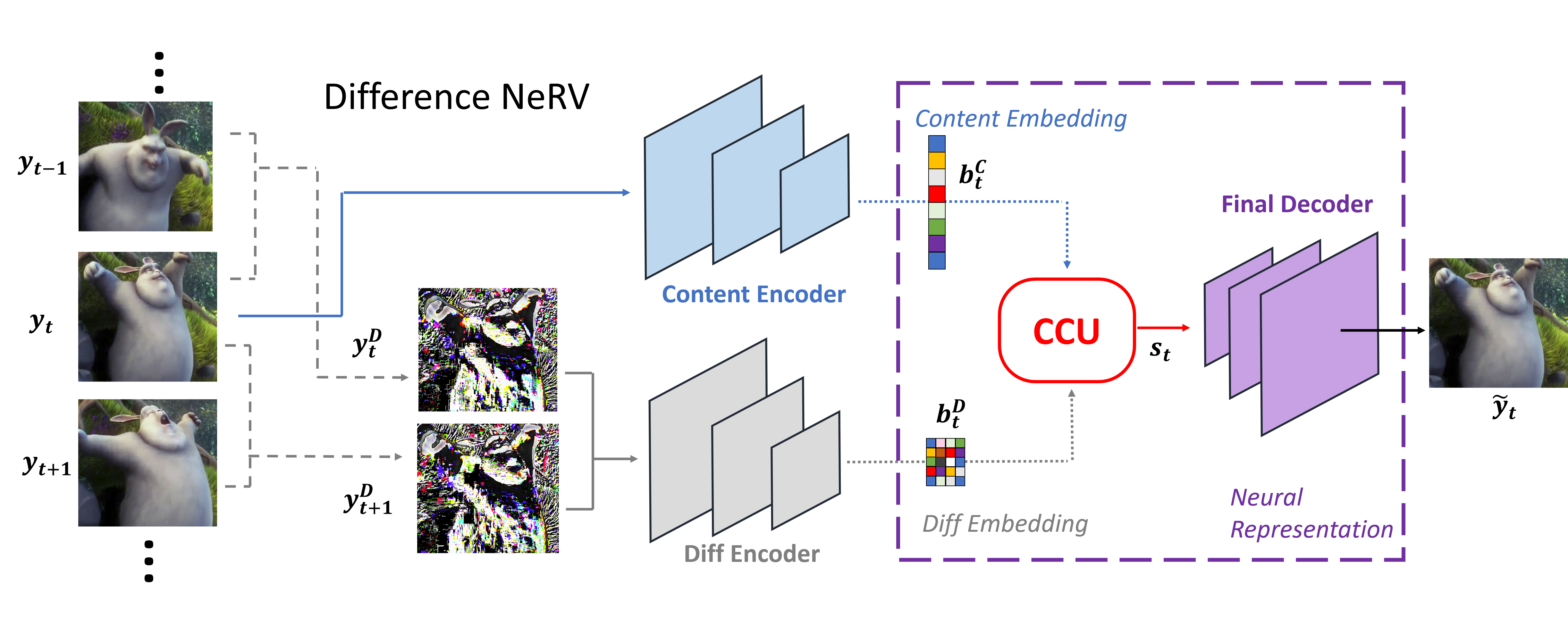}
  \caption{The pipeline of DNeRV. Blue part indicates content stream, grey for diff stream and red for fusion. The purple part is the implicit neural representation for the given video, consisting of embeddings, lightweight CCU, and decoder.} \label{dnerv}
\end{figure*}

In this work, we propose {Difference NeRV} (DNeRV) that attempts to approximate a \textit{dynamical system} by absorbing the difference of adjacent frames, $\mathbf{y}^{D}_t=\mathbf{y}_t - \mathbf{y}_{t-1}$ and $\mathbf{y}^{D}_{t+1}=\mathbf{y}_{t+1} - \mathbf{y}_{t}$, as a {diff stream} input. Further analysis for the importance of diff stream is presented in Section~\ref{mot}. An illustration of DNeRV pipeline is presented in Figure~\ref{dnerv}. Diff encoder captures short-term contextual correlation in the diff stream, which is then merged with the content stream for spatiotemporal feature fusion. In addition, we propose a novel gated mechanism, collaborative content unit (CCU), which integrates spatial features in the content stream and temporal features in the diff stream to obtain accurate reconstruction for those frames with adjacent dynamics.

The main contribution of this paper are as follows.
\begin{itemize}[leftmargin=4mm,topsep=0pt,itemsep=0ex,partopsep=1ex,parsep=1pt]
   \item Existing NeRV methods cannot model content-specific features and contextual correlations simultaneously. We offer an explanation using adjacent dynamics. Furthermore, we reveal the importance of diff stream through heuristic analysis and experiments.
   \item We propose the Difference NeRV, which can model the content-specific spatial features with short-term temporal dependence more effectively and help network fit the implicit mapping efficiently. We also propose a collaborative content unit to merge the features from two streams adaptively.
   \item We present experiments on three datasets (Bunny, UVG, and Davis Dynamic) and various downstream tasks to demonstrate the effectiveness of the proposed method. The superior performance over all other implicit methods shows the efficacy of modeling videos with large motion. As a result, DNeRV can be regarded as a new baseline for INR-based video representation.
\end{itemize}

\section{Related Work}
\noindent \textbf{Implicit neural representations (INRs)} have been used in various vision tasks in recent years \cite{DBLP:conf/cvpr/ChenZ19, nerf}. In 3D vision, \cite{rawnerf, DeblurNeRF, DNeRF, pointNeRF, pointNeRF} aim to use INRs from static and simple to dynamic and complicated visual data. In image analysis, INRs have been used to learn the mapping between 2D spatial coordinates $\mathbf{x} \in \mathbb{R}^2$ and corresponding RGB value $\mathbf{y} \in \mathbb{R}^3$ via various positional embedding techniques~\cite{siren, fourier, mfn,DBLP:conf/cvpr/YuceOBF22} or meta-learning~\cite{Strmpler2022ImplicitNR}. In video analysis, INRs learn the mapping from frame index $x_{t} \in \mathbb{R}$ to RGB frame $\mathbf{y} \in \mathbb{R}^{3 \times w \times h}$~\cite{nerv, VideoINR, DBLP:conf/cvpr/Mai022}. In other visual tasks,  INRs can encode both spatial coordinates and specific feature vectors~\cite{DBLP:journals/corr/abs-2203-00137, DBLP:conf/iclr/YuTMKK0S22}.

\noindent \textbf{Neural representation for videos (NeRV)} methods can be broadly divided into two groups. \textit{Index-based} NeRV methods use the positional embedding of $t$ as the input~\cite{nerv, e-nerv, DBLP:conf/cvpr/Mai022}. The follow-up work has improved network structure for acceleration or disentangled feature modeling, but those methods could not capture the content-specific information, causing spatial redundancy. Hybrid-based NeRV \cite{HNeRV22} provides an insightful view by treating the current frame itself as the index embedding. The method shows much better performance over index-based ones via content-adaptive fashion. The main limitation of hybrid NeRV is that they ignore temporal relationship between frames, resulting in poor performance with adjacent dynamics.

\noindent \textbf{Video compression} methods based on INR use the traditional pipeline but change some intermediate components into networks~\cite{dvc, FVC,MLVC,DBLP:conf/nips/LiLL21,ELFVC}. However, their utility is limited by large number of parameters and computational redundancy, causing long coding time and limited generalized ability. One significance of INRs for video is that the video compression can be regarded as a model compression problem, and the existing model compression and acceleration techniques~\cite{Gholami2018SqueezeNextHN, Frankle2019TheLT} could be used for INR-based network. Neural compression is expected to break through the limitation of traditional pipeline with the help of INRs, resulting in better performance.

\noindent \textbf{Two stream vision and fusion} resemble the idea of adopting multi-stream mechanism for disentangling feature learning in video analysis. \cite{DBLP:conf/dagm/Sevilla-LaraLGJ18} observe the importance of optical flow, which is widely used in action recognition~\cite{DBLP:conf/nips/SimonyanZ14, DBLP:conf/cvpr/FeichtenhoferPZ16, I3D} or other visual tasks~\cite{DBLP:conf/cvpr/SunY0K18, DBLP:conf/eccv/HuangZHSZ22}. Separating motion and content as two stream is also widely used in video prediction~\cite{DBLP:conf/iclr/VillegasYHLL17} and image-to-video generation~\cite{DBLP:conf/eccv/ZhaoPTKM18}. Furthermore, hidden state and time-dependent input in RNN/LSTM could be thought as different stream~\cite{DBLP:journals/neco/HochreiterS97, DBLP:journals/corr/ChungGCB14, DBLP:conf/nips/WangLWGY17}. More effective fusion module for different stream information is a fundamental task in sequence learning, generating lots of novel gated mechanism~\cite{DBLP:conf/icml/WangGLWY18, DBLP:conf/cvpr/ZhangDLH018, DBLP:conf/icml/GuGP0P20}.
The two-stream fusion aims to reduce the \textit{spatiotemporal redundancy} caused by the inherent continuity of video, which motivated us to propose DNeRV based on diff stream. Additionally, inspired by the gated mechanism in RNN, we introduce a fusion unit for adaptive feature fusion.

\section{Motivation}~\label{mot}

\begin{figure*}[t]
  \centering
  \includegraphics[width=0.8\linewidth]{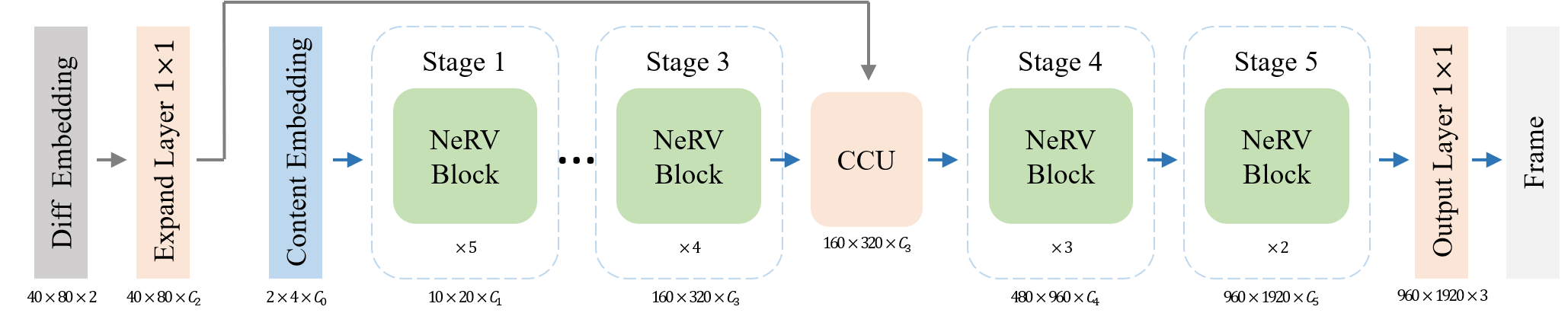}\\
  \caption{Architecture of decoder with CCU as fusion module for $960 \times 1920$.}\label{dec}
\end{figure*}

In this section, we discuss how the frame differences benefit the adjacent dynamics.
A video sequence is a collection of images captured over time; therefore, motion of different objects in video is often spatially continuous and smooth. The main objective of INR is to learn an implicit mapping $f(\mathbf{x}) \mapsto \mathbf{y}$. Thus, we usually assume that training tuples $(\mathbf{x},\mathbf{y})$ are dense enough so that the continuous $f$ can be approximated. In the case of NeRV, the difference between frames (or the so-called adjacent dynamics) violates the expected smoothness. That is the main reason for the failure of NeRV in the presence of large motion and highly dynamic scenes.

For video INRs, the training data are $\{(t, \mathbf{y}_t)\}_{t=0}^N$, where $\mathbf{y}_t$ is the frame at time index $t$. We assume that the implicit (unknown) mapping $f$ is continuous and defined as
\begin{equation}\label{f}
f: \Phi \rightarrow \mathbb{R}^{3 \times H \times W}, \quad  \Phi = [0,N].
\end{equation}
Let us note that although NeRV methods use various positional encoding techniques \cite{fourier, siren} to map the discrete index into a higher dimension for better performance (we treat the frame itself as an embedding of the index for hybrid-based methods, the network $g_{\theta}$ is actually a mapping from a time index in $\Phi$ to $\mathbb{R}^{3 \times H \times W}$. Once trained, we can generate frames as
\begin{equation}\label{trained_g}
 g_{\theta} \left( t \right) =  \mathbf{y}_t, \quad t \in \{1,2,\cdots,N\}.
\end{equation}

Since video is usually continuous along spatial and temporal dimensions, $f$ can also be described as a dynamical system:
\[
\dot f = A(f(t),t),
\]
where $A(\cdot,\cdot)$ represents a nonlinear function determined by the system. The goal of INR is to use the neural network $g_{\theta}$ to fit the $f$ with the training tuples $\{(t, f(t))\}_{t=0}^N$.
However, in general, the training tuple $(t, f(t))$ is not enough for networks to learn a dynamical system. For example, we consider a simple time-invariant system:
\[
\dot f = A \cdot f(t),\quad f(0) = \mathbf{y}_0,
\]
where $A$ is a constant. The solution of the problem can be  written as $f(t) = \exp(At)\mathbf{y}_0$.
If $g_{\theta}$ is a general feed-forward ReLU network, then we can achieve $\|g_{\theta}(t) - f(t)\|\leq \epsilon$ for all $t$ under some conditions. In particular, we require $O(\log(\epsilon)^{d})$ non-zero hidden units to train $g_{\theta}(t)$ using training tuples $\{(t, f(t))\}$, where $d$ is the dimension of $f(t)$ if $f$ is invertible\cite{SchmidtHieber2020NonparametricRU, MCN}.

Interestingly, if we use $\{(t, \dot f(t)),f(t)\}$  as training tuples, then one-layer linear network is sufficient to learn $f(t)$ and
\[
\|g_{\theta}(t, \dot f(t)) - f(t)\| = 0, \quad \forall
 t \in [1,N].
\]
This is because the learning problem simplifies to learning the constant $A$ instead of learning a continuous function $f(\cdot)$. Hence, considering the high order differentials $(\dot f, \ddot f, \cdots)$ as the network input can significantly benefit in learning a dynamical system.

For hybrid-based INR methods, we introduce the \textit{diff stream} $\mathbf{y}^{D}_t=\mathbf{y}_t - \mathbf{y}_{t-1}$ as an auxiliary input with the main content stream $\mathbf{y}_t$ (for more details, see the  ablation studies). The difference between two adjacent frames could be treated as discrete differential $\nabla f $,
\begin{equation}\label{diff}
\begin{split}
\nabla f \bigg|_{\tau = t}  &= \frac{f(\tau)-f(\tau-\Delta \tau)}{\Delta \tau}\bigg|_{\tau = t} \\
 & \simeq \frac{f(t)-f(t-1)}{t-(t-1)}  = \mathbf{y}_t - \mathbf{y}_{t-1}.
\end{split}
\end{equation}
In addition to more efficiently fitting the implicit function $f$, we believe that diff stream could also help $g_\theta$ capture the adjacent dynamic and a more robust video representation.

\section{Method}
\noindent \textbf{Overview.} Our proposed DNeRV method is a hybrid-based approach that uses two input streams. The encoders in DNeRV process two streams separately; the intermediate output embeddings pass through a fusion step before entering the decoders. We follow the \textit{hybrid} philosophy in~\cite{HNeRV22} that treats the \textit{video-specific embeddings} and \textit{video-specific decoder} as the INRs for videos. The model architecture is shown in Fig. \ref{dnerv}.

\noindent \textbf{Encoder.} The encoder consists of Diff Encoder and Content Encoder. For the content encoder, we use the settings in \cite{HNeRV22} for a fair comparison, and use input as the current frame $\mathbf{y}_t$. Meanwhile, for the given frame $\mathbf{y}_t$, DNeRV calculates the frame differences,
$\mathbf{y}^{D}_t=\mathbf{y}_t - \mathbf{y}_{t-1}$ and $\mathbf{y}^{D}_{t+1}=\mathbf{y}_{t+1} - \mathbf{y}_{t}$, and concatenates them before feeding into diff encoder. Both content encoder and diff encoder adopt one ConvNeXT block at one stage, with a scale transformation layer and layer norm adopted at the beginning of each stage. For $1920 \times 960$ video, the strides are [5, 4, 4, 3, 2] in content encoder and [4, 3, 2] in diff encoder, providing content embedding $\mathbf{b}^{C}_{t}$ in $16 \times 2 \times 4$ and diff embedding $\mathbf{b}^{D}_{t}$ in $2 \times 40 \times 80$. It is worth mentioning that the shape of diff embedding is flexible, smaller diff embedding (e.g., $2 \times 10 \times 20$) only brings slight performance drops and remain comparable against other NeRV methods. See more details in ablation studies and Tab.~\ref{fusion}. The encoder can be described as
\begin{equation}\label{enc}
\begin{split}
\mathbf{b}^{C}_{t} &= \mathsf{CONT\_ENC}( \mathbf{y}_{t}), \\
\mathbf{b}^{D}_{t} &= \mathsf{DIFF\_ENC}( \mathsf{concat}[\mathbf{y}^{D}_{t}, \mathbf{y}^{D}_{t+1}]).
\end{split}
\end{equation}

\noindent \textbf{Decoder.} We adopt NeRV block as the basic block in each decoding stage. For the embeddings from encoder, they would be fused in same shape. We explored different connections for fusion, and the final architecture is shown in Fig. \ref{dec}. For the fusion module, it could be sum fusion, conv fusion, or other gated mechanisms. Once the features are merged, they pass through other stages and map into pixel domain via channel reduction.

\noindent \textbf{Fusion.} To fuse features from diff stream and content stream, conv fusion $\mathbf{s}_{t}= \mathsf{Conv}(\mathbf{b}^{C}_{t}) + \mathbf{b}^{D}_{t}$ or concat fusion $\mathbf{s}_{t}=  \mathsf{concat}[ \mathbf{b}^{C}_{t}, \mathbf{b}^{D}_{t}]$ may not be suitable. This is because the features come from different domains as $\mathbf{b}^{D}$ is the discretization of differential of $f$, while $\mathbf{b}^{C}$ represents the value of $f$. To merge the two streams, inspired by gated mechanism in temporal sequence learning~\cite{DBLP:journals/corr/ChungGCB14, DBLP:conf/iclr/YuLEF20, DBLP:conf/icml/WangGLWY18}, we introduce a {collaborative content unit} (CCU).
Our motivation is that diff stream needs to be fused with \textit{content} stream \textit{collaboratively} to obtain the refined content features by adding high-order information. Specifically, the content stream could be treated as hidden state in an RNN-fashion, which contains time-varying features helping reconstruction. CCU can be represented as
\begin{equation}\label{ccu}
\begin{split}
\mathbf{z}^{D}_{t}&= GELU( \mathsf{PS}(\mathbf{W}^{1 \times 1}_{z} \ast (\mathbf{b}^{D}_{t})) ),\\
\mathbf{\tilde{b}}^{C}_{t}&= \mathsf{BLOCK}^{(2)}(\mathbf{b}^{C}_{t}), \\
\mathbf{u}_{t}&= tanh (\mathbf{W}_{ub} \ast \mathbf{\tilde{b}}^{C}_{t} + \mathbf{W}_{uz} \ast \mathbf{z}^{D}_{t} ), \\
\mathbf{v}_{t}&= Sigmoid (\mathbf{W}_{vb} \ast\mathbf{\tilde{b}}^{C}_{t} + \mathbf{W}_{vz} \ast \mathbf{z}^{D}_{t}), \\
\mathbf{s}_{t} &=  \mathbf{u}_{t} \odot \mathbf{v}_{t} + (1-\mathbf{v}_{t}) \odot \mathbf{\tilde{b}}^{C}_{t},
\end{split}
\end{equation}
where $\mathsf{PS}$ is PixelShuffle, $\ast$ is convolution operator and $\odot$ is Hadamard product. $\mathbf{v}_{t}$ could be treated as the update gate in GRU~\cite{DBLP:journals/corr/ChungGCB14}, to decide how much information in content feature could be remained. Finally, two streams are merged and the adjacent dynamics collaboratively captured by CCU can help network $g_{\theta}$ learn the implicit mapping. To balance the parameter quantity, we reduce the channels in the last two stages of the decoder. Final output is given as follows
\begin{equation}
\begin{split}
\mathbf{s}_{t} &= \mathsf{FUSION}( \mathbf{b}^{C}_{t}, \mathbf{b}^{D}_{t}), \\
\mathbf{\tilde{y}}_{t} &= Sigmoid ( \mathbf{W}^{1 \times 1}_{y} \ast  ( \mathsf{BLOCK}^{(3)}( \mathbf{s}_{t} ) ) ),
\end{split}
\end{equation}
where $\mathsf{FUSION}$ represents CCU in our implementation.

\begin{table*}[t]
    \label{Tab}
    \begin{subtable}{.5\linewidth}
      \centering
       \tabcolsep=0.1cm
        \resizebox{!}{1.0cm}{
        \begin{tabular}{l|cccc}
     size & 0.35M & 0.75M& 1.5M& 3M \\
    \midrule[1.5pt]
    NeRV~\cite{nerv} & 26.99 &  28.46 & 30.87 & 33.21  \\
    E-NeRV~\cite{e-nerv} & 27.84&  30.95& 32.09& 36.72	 \\
    H-NeRV~\cite{HNeRV22} &30.15& 32.81& 35.19& 37.43 \\
    \midrule
    \textbf{D-NeRV} & \textbf{30.80}	&\textbf{33.30}	&\textbf{35.22 }	&\textbf{38.09}\\
        \end{tabular}
        }\setlength{\belowcaptionskip}{5pt}
        \caption{PSNR on Bunny with varying \textbf{model size}.}\label{vr_size}
    \end{subtable}%
       \begin{subtable}{.5\linewidth}
      \centering
       \tabcolsep=0.1cm
        \resizebox{!}{1.0cm}{
        \begin{tabular}{l|cccccc}
     epochs & 300 & 600& 1200 & 1800 &2400 &3600 \\
    \midrule[1.5pt]
    NeRV~\cite{nerv}          & 28.46 &  29.15 & 29.57 & 29.73 & 29.77 & 29.86  \\
    E-NeRV~\cite{e-nerv} & 30.95&  32.07  & 32.79& 33.10	 & 33.36& 33.67\\
    H-NeRV~\cite{HNeRV22} &32.81& 33.89& 34.51& 34.73 & 34.88& 35.03\\
     \midrule
    \textbf{D-NeRV} & \textbf{33.30}&\textbf{34.28}	&\textbf{34.83 }&\textbf{35.16} &\textbf{35.25 }	&\textbf{35.34}\\
       \end{tabular}
        }\setlength{\belowcaptionskip}{5pt}
        \caption{PSNR on Bunny with varying \textbf{epochs}.}\label{vr_epoch}
    \end{subtable}
    \\
    \begin{subtable}{.5\linewidth}
      \centering
       \tabcolsep=0.05cm
        \resizebox{!}{1.1cm}{
        \begin{tabular}{l|ccccccc|c}
     960$\times$1920 & Beaut & Bosph & Honey & Jocke & Ready & Shake &Yacht &avg.\\
    \midrule[1.5pt]
    NeRV~\cite{nerv} & 33.25 & 33.22 & 37.26    &31.74   & 	24.84 & 33.08   & 28.03 &31.63  \\
    E-NeRV~\cite{e-nerv} &33.17 & 33.69 & 37.63    &31.63   & 	25.24  &34.39 	& 28.42  &32.02 \\
    HNeRV~\cite{HNeRV22} &33.58 & 34.73 & 38.96    &32.04 	  & 25.74  &34.57 	& 29.26 & 32.69\\
 \midrule
 \textbf{DNeRV} (L1+SSIM) &\textbf{40.19}  & 36.59  & \textbf{43.23}  &35.75 & 28.17   &\textbf{38.25}  & 30.73 & \textbf{36.13} \\
  \textbf{DNeRV} (L2)& 40.00 &\textbf{36.67} &41.92  &\textbf{35.75} &\textbf{28.67} &36.53 &\textbf{31.10} &35.80\\
         \end{tabular}
        }\setlength{\belowcaptionskip}{-3pt}
        \caption{PSNR on UVG in  960 $\times$ 1920.}\label{vr_uvg_960}
    \end{subtable}%
       \begin{subtable}{.5\linewidth}
      \centering
       \tabcolsep=0.05cm
        \resizebox{!}{1.1cm}{
        \begin{tabular}{l|ccccccc|c}
     480$\times$960 & Beaut & Bosph & Honey & Jocke & Ready & Shake &Yacht &avg.\\
    \midrule[1.5pt]
    NeRV~\cite{nerv} & 36.27   & 35.07 & 40.76   &32.58    &25.81   & 35.33    & 30.11 & 33.70 \\
    E-NeRV~\cite{e-nerv} &36.26 & 36.06 & 43.26   &32.70     & 26.19     &35.64   & 30.38 & 34.35\\
    HNeRV~\cite{HNeRV22} &36.91 & 36.95 & 42.05  &33.33 & 27.07  &36.97  & 30.96 & 34.89 \\
   \midrule
     \textbf{DNeRV} (L1+SSIM) &\textbf{40.24} & 37.35  & \textbf{43.98}  &\textbf{35.85} & 28.70  &\textbf{38.84}  & 31.03 & \textbf{36.58} \\
    \textbf{DNeRV} (L2) & 39.64	&\textbf{37.49}	&42.45 &35.44 	&\textbf{29.21}	&36.83 &\textbf{31.30} &36.05\\
       \end{tabular}
        }\setlength{\belowcaptionskip}{-3pt}
        \caption{PSNR on UVG in  480 $\times$ 960.}\label{vr_uvg_480}
    \end{subtable}
    \caption{Video regression results on Bunny and UVG, where DNeRV uses different loss functions for ablation.  }
  \end{table*}

\noindent \textbf{Discussion of optical flow.}
Although optical flow captures adjacent temporal relationship as well as the difference stream, we could not achieve comparable performance when using optical flow. The main reason is that INR-based video representation task is different from semantic video tasks. In the case of NeRV, pixel-level features that directly help decoder reconstruction are more vital. More details can be found in the supplementary materials.

\noindent \textbf{Comparison with NeRV.}
Now, we look back to the philosophy of NeRV and compare it with DNeRV. For NeRV, we search for an operator $g_{\theta}$ by solving the following optimization problem:
\begin{equation}\label{g_search}
\begin{split}
 \mathop{\text{argmin}}\limits_{\theta} \Vert  g_{\theta} \left( h( t ) \right)- f(t)  \Vert, \\
\end{split}
\end{equation}
where $h$ represents the embedding of time index $t$ and $f(t)=\mathbf{y}$ represents the frames in pixel-domain. In the case of hybrid methods~\cite{HNeRV22}, we solve the following optimization problem:
\begin{equation}\label{g_search}
\begin{split}
 \mathop{\text{argmin}}\limits_{\theta} \Vert  g_{\theta} \left( f( t ) \right)- f( t )  \Vert, \\
\end{split}
\end{equation}
where the embedding is the frame itself. The hybrid method attempts to fit a series of invariant point transformations in function space for every training tuple $(t, \mathbf{y})$. This explains why existing methods only work well on fixed background scene with few dynamics, such as ``HoneyBee'' or ``ShakeNDry'' in UVG. In other words, they only take effect when $\mathbf{y}_i$ is within a small neighborhood of training samples.
%
In other words, $g_\theta$ only learns the behavior of $f$ near the mean of whole training samples, where adjacent dynamics would not be apparent.
In the case of DNeRV, we solve the following optimization problem:
\begin{equation}\label{g_search}\vspace{-0.05in}
\begin{split}
 \mathop{\text{argmin}}\limits_{\theta} \Vert  g_{\theta} \left( f, \nabla f, ..., \nabla^{(i)} f\right)- f  \Vert, \\
\end{split}
\end{equation}
where $i=1$ in our realization. DNeRV attempts to learn a \textit{dynamical system} that represents $f$ in implicit way. 

\section{Experiments}
\noindent \textbf{Settings}. We verify DNeRV on Bunny~\cite{bunny}, UVG~\cite{uvg} and DAVIS Dynamic. Bunny owns 132 frames for $720 \times 1280$. UVG has 7 videos at $1080 \times 1920$ with length of 600 or 300. DAVIS Dynamic is a subset of DAVIS16 validation~\cite{davis} which containing 22 videos\footnote{blackswan, bmx-bumps, camel, breakdance, car-roundabout, bmx-trees, car-shadow, cows, dance-twirl, dog, car-turn, dog-agility, drift-straight, drift-turn, goat, libby, mallard-fly, mallard-water, parkour, rollerblade, scooter-black, strolle.} in $1080 \times 1920$. Most of the selected videos contain dynamic scenes or moving targets, which are quite difficult for existing methods. Following the settings in \cite{HNeRV22} for fair comparison, we center-crop the videos into $640 \times 1280$ or $960 \times 1920$ and reshape UVG into $480 \times 960$ for additional comparison.

During training, we adopt Adam~\cite{adam} as the optimizer with learning rate of $5\times 10^{-4}$ and cosine annealing learning rate schedule~\cite{Wang2003MultiscaleSS} and the batch size is set to 1. We use PSNR and SSIM to evaluate the video quality. The stride list, kernel size and reduction rate remain to be same as \cite{HNeRV22}, except for the channels in the last two stages of decoder.

We compare DNeRV with others in video regression and three downstream visual tasks consist of video compression, interpolation and inpainting. In video interpolation we train the model on the sequence of even frames and test it on the odd frames sequence from UVG and DAVIS Dynamics. In video inpainting, we directly use the models trained in regression without any fine-tuning and test them using masked videos in disperse mask or central mask from DAVIS Dynamics. All experiments are conducted in Pytorch with GPU RTX2080ti, with 3M size and 300 epochs unless otherwise clarified.

\noindent \textbf{Discussion of Loss functions}. We conduct loss objective ablation between L2 and L1+SSIM, shown in Tab.~\ref{vr_uvg_960} and Tab.~\ref{vr_uvg_480}. L1+SSIM is the loss objective in NeRV~\cite{nerv}, $L(\mathbf{\hat{y}}, \mathbf{y}) = \alpha \Vert \mathbf{\hat{y}}, \mathbf{y} \Vert_1  + (1-\alpha)(1-SSIM(\mathbf{\hat{y}}, \mathbf{y})), \alpha=0.7$. Owing to L1 norm is the convex approximation of L0 norm~\cite{Cands2005DecodingBL}, it is better for scenes with complex textures and high-frequency subtle spatial structure but few motion between frames. While during experiments, L2 is set as default loss function as it is better for low-frequency scenes with large motion.

\subsection{Video Regression}
\noindent \textbf{Bunny.} The comparison between different implicit methods trained in 300 epochs on Bunny is shown in Tab.~\ref{vr_size}. DNeRV outperforms others. Also, we compare various implicit methods in same 0.75M but different training epochs reported in Tab.~\ref{vr_epoch}. DNeRV surpasses other methods relying on the reasonable structure with no additional gradient propagation difficulties.

\noindent \textbf{UVG.} The PSNR results on UVG are given in Tab.~\ref{vr_uvg_960} and Tab.~\ref{vr_uvg_480}. DNeRV shows large improvements at resolution $960 \times 1920$. The excellent results are attributed to the high-resolution diff stream, containing a great deal of content-specific spatial information. The adjacent dynamic hidden among frames could be captured in a more reasonable way, helping the network converge faster.

\noindent \textbf{DAVIS Dynamic}. The results of regression on DAVIS Dynamic are shown together with inpainting results in Tab.~\ref{vr_davis}. DNeRV achieves an impressive performance when processing those videos with complicated scene transformation or object movements. Another difficulty of DAVIS Dynamic videos is that the number of frames is quite smaller, e.g., 25 for "dog-agility" and 43 for "Scooter-black". Fewer frames and ubiquitous adjacent dynamics present extreme difficulty for implicit methods to fit the latent mapping, indicating the effectiveness of DNeRV.

\begin{figure}[t]
  \centering
  \includegraphics[width=1\linewidth]{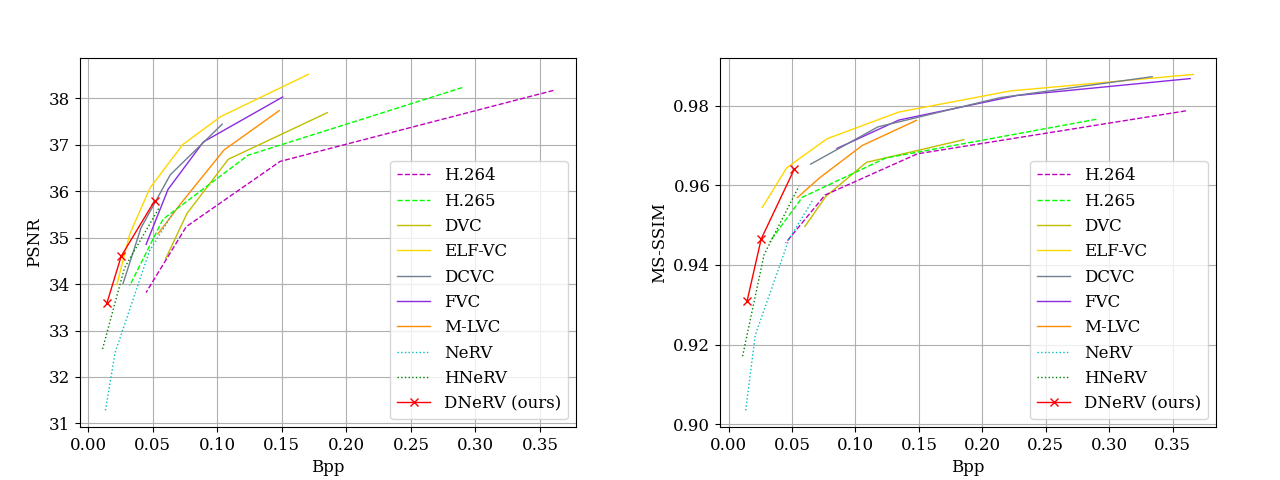}\\
  \caption{Compression results on 960 $\times$ 1920 UVG. DNeRV outperforms other INR-based methods.} 
  \label{uvg}
\end{figure}

\begin{table*}[t]\footnotesize
 \setlength{\tabcolsep}{3pt}
  \centering
  \begin{tabular}{l||cccc||cc||cc}
 \toprule[1.5pt]
 \multirow{2}*{\shortstack{video}} &\multicolumn{4}{c||}{regression}  &\multicolumn{2}{c}{Inp-Mask-S}  &\multicolumn{2}{c}{Inp-Mask-C} \\
      & NeRV & E-NeRV & HNeRV & DNeRV & HNeRV & DNeRV & HNeRV & DNeRV \\
    \midrule
    Blackswan &28.48/0.812 &29.38/0.867  &30.35/0.891 &\textbf{30.92/0.913}  &26.51/0.825 &\textbf{29.01/0.886}
   &24.64/0.783 &\textbf{27.45/0.858}\\
   Bmx-bumps  &29.42/0.864 &28.90/0.851  &29.98/0.872 &\textbf{30.59/0.890}  &23.16/0.728 &\textbf{25.70/0.819}
   	&20.39/0.665 &\textbf{22.95/0.767}\\
   Bmx-trees  &26.24/0.789 &27.26/0.876  &28.76/0.861 &\textbf{29.63/0.882}  &22.93/0.720 &\textbf{26.57/0.841}
   &20.26/0.653 &\textbf{21.62/0.752}	\\
   Breakdance &26.45/0.915 &28.33/0.941  &30.45/0.961 &\textbf{30.88/0.968}  &27.63/0.945 &\textbf{29.16/0.961}
  &23.84/0.907 &\textbf{25.18/0.938}\\
    Camel	  &24.81/0.781 &25.85/0.844  &26.71/0.844 &\textbf{27.38/0.887}  &20.94/0.661 &\textbf{24.71/0.832}
   	&21.85/0.733 &\textbf{23.72/0.815}\\
   Car-round  &24.68/0.857 &26.01/0.912  &27.75/0.912 &\textbf{29.35/0.937}  &23.96/0.752 &\textbf{27.63/0.854}
   	&22.06/0.810 &\textbf{24.82/0.886}\\
   Car-shadow &26.41/0.871 &30.41/0.922  &31.32/0.936 &\textbf{31.95/0.944}  &25.87/0.875 &\textbf{27.90/0.914}
  	&\textbf{28.67}/0.908 &28.18/\textbf{0.923} \\
   Car-turn   &27.45/0.813 &29.02/0.888  &29.65/0.879 &\textbf{30.25/0.892}  &23.96/0.752 &\textbf{27.63/0.854}
   	&24.43/0.773 &\textbf{25.67/0.821}\\
   Cows       &22.55/0.702 &23.74/0.819  &24.11/0.792 &\textbf{24.88/0.827}  &21.37/0.682 &\textbf{22.91/0.770}
   &20.81/0.668 &\textbf{21.87/0.733}	\\
   Dance-twril&25.79/0.797 &27.07/0.864  &28.19/0.845 &\textbf{29.13/0.870}  &23.05/0.743 &\textbf{26.13/0.830}
  &21.10/0.704 &\textbf{23.00/0.783}	\\
   Dog        &28.17/0.795 &30.40/0.882  &30.96/0.898 &\textbf{31.32/0.905}  &25.34/0.739 &\textbf{27.43/0.824}
  &23.09/0.677 &\textbf{24.96/0.763}	\\
    Dog-ag     &29.08/0.821 &29.30/0.905  &28.75/0.893 &\textbf{29.94/0.923}  &27.70/0.884 &\textbf{28.28/0.913}
    &\textbf{25.12}/0.856 &24.85/\textbf{0.886}\\
   Drift-straight&26.65/0.860 &29.10/0.941 &30.80/0.932 &\textbf{31.50/0.940}&24.79/0.833 &\textbf{27.00/0.892}
  &20.15/0.725 &\textbf{22.61/0.823}	 \\
   Drift-turn &26.70/0.812 &27.94/0.875  &29.72/0.834 &\textbf{30.37/0.862}  &22.27/0.677 &\textbf{26.20/0.816}
   &19.95/0.636 &\textbf{22.65/0.749}	\\
   Goat       &23.90/0.746 &25.25/0.855  &26.62/0.858 &\textbf{27.79/0.887}  &21.11/0.675 &\textbf{22.65/0.755}
   &20.21/0.639 &\textbf{21.76/0.727}\\
   Libby      &29.08/0.821 &31.43/0.890  &32.69/0.917 &\textbf{33.43/0.927}  &27.59/0.825 &\textbf{29.51/0.875}
  &24.33/0.752 &\textbf{26.07/0.826}	\\
   Mallard-fly &26.83/0.757 &28.84/0.847 &\textbf{29.22/0.848} &28.77/0.833  &23.81/0.709 &\textbf{25.83/0.774}
  &20.83/0.618 &\textbf{22.77/0.689}	 \\
   Mallard-water&25.20/0.824 &27.28/0.896 &29.08/0.908&\textbf{29.69/0.922}  &23.55/0.803 &\textbf{24.11/0.845}
  &\textbf{21.18}/0.743 &21.15/\textbf{0.796}	 \\
   Parkour	  &25.14/0.794 &25.31/0.845  &\textbf{26.56/0.851} &\underline{25.75/0.827}  &21.32/0.685 &\textbf{24.51/0.799}
   &19.97/0.650 &\textbf{21.55/0.754}	\\
   Rollerblade &29.28/0.898 &\textbf{33.32/0.964} &32.19/0.935 &\underline{32.49/0.940}  &29.32/0.911 &\textbf{30.41/0.931}
  &27.31/0.901 &\textbf{27.63/0.917}	\\
   Scooter-black&22.73/0.835&25.79/0.927 &27.38/0.923 &\textbf{28.53/0.940}  &21.05/0.794 &\textbf{24.27/0.897}
  &19.76/0.789   &\textbf{21.00/0.844}	\\
   Stroller    &29.28/0.859  &30.37/0.914 &31.31 0.894&\textbf{32.73/0.928}  &25.90/0.796 &\textbf{28.46/0.876}
  &22.86/0.734  &\textbf{24.52/0.822}	\\
  \midrule
   Average     &26.56/0.819 &28.19/0.887 &29.21/0.886 &\textbf{29.86/0.902} &24.09/0.774  &\textbf{26.53/0.854}
   	&22.40/0.742 &\textbf{23.91/0.812} \\
   \bottomrule[1.5pt]
  \end{tabular}
  \caption{Video regression and inpainting results on 960$\times$1920 Davis Dynamic in PSNR/SSIM, larger is better.}
  \label{vr_davis}
\end{table*}

\begin{table}[h]\footnotesize
  \centering
  \begin{tabular}{c|ccc|c|c}
    \multirow{2}*{\shortstack{size of \\ diff embedding }} &\multicolumn{3}{c|}{stage} &\multicolumn{1}{c|}{params}  &\multicolumn{1}{c}{PSNR/SSIM}    \\
      &2 &3& 4  &&\\
     \midrule[1.5pt]
      N/A &&& &0.311M   &29.64/0.908\\
          &&& &0.347M   &29.95/0.914\\
    \midrule
    $2 \times 40 \times 80$ &A &&  &0.288M   &29.31/0.907\\
     &A &C&     &0.339M   &29.93/0.914\\
     & &\underline{C}&  &0.339M      &\underline{30.60/0.924}\\
     &A &&C    &0.348M    &29.73/0.909\\
     & &&C     &0.348M    &30.38/0.922\\
     & &C&C   &0.348M     &30.33/0.921\\
     &A &C&C  &0.348M     &26.96/0.835\\
 \midrule
   $2 \times 10 \times 20$ & &U& &0.343M  &29.66/0.907\\
  \midrule
    \textbf{Final}&&U& &0.349M &\textbf{30.80/0.930} \\
  \end{tabular}
  \caption{Ablation study for fusion module and diff embedding size, training on Bunny in 300 epochs. \textbf{A} indicates sum fusion, \textbf{C} is conv fusion and \textbf{U} is the CCU. The final version of DNeRV consists of diff embedding in shape of $2 \times 40 \times 80$, 3rd stage where merging and CCU as the fusion module. The first two rows which are marked as N/A represent HNeRV baseline, where the size of diff embedding is not available.}
  \label{fusion}
\end{table}

\begin{table}[t]\footnotesize
  \centering
  \tabcolsep=0.1cm
  \begin{tabular}{l|c|c}
   input of diff stream   &conv fusion &CCU \\
    \midrule[1.5pt]
    $\Delta f(t)$                           &30.359/0.920  &30.425/0.922 \\
    $\Delta f(t+1)$                         &30.354/0.919  &-/-\\
    $(\Delta f(t-1) + \Delta f(t+1))/2$     &30.213/0.918  &30.278/0.920 \\
    $\mathsf{concat}[ \Delta f(t),\Delta f(t+1)]$  &\underline{30.598/0.924} &\textbf{30.804/0.930} \\
$\mathsf{concat}[\Delta f(t), \Delta f(t+1), \nabla f(t)]$&30.310/0.919  &30.392/0.921\\
  \end{tabular}
  \caption{Ablation study of various difference on Bunny in 0.35M and 300 epochs, where $\Delta$ is first order and $\nabla$ is second order difference.}
  \label{diff_abla}
\end{table}

\subsection{Video Compression}
We show the compression results in PSNR and SSIM on UVG dataset in Fig.~\ref{uvg}. Without pruning and only 8 bits quantization with entropy encoding adopted, DNeRV outperforms other NeRV methods especially on PSNR. DNeRV optimizes the network structure under the condition of parameter amount, thus reducing the redundancy in weights. Although it couldn't maintain performance in 40\% pruning like~\cite{nerv, e-nerv}, the 10\% pruned DNeRV is still competitive compared with other implicit methods. We also report VMAF~\cite{Rassool2017VMAFRV} results on UVG in the appendix.

\begin{table}[h]\footnotesize
  \centering
  \begin{tabular}{l|lrrl}
     method & 1920  &params$\downarrow$ & dec time$\downarrow$ & FPS$\uparrow$ \\
    \midrule[1.5pt]
    DCVC~\cite{DBLP:conf/nips/LiLL21} &$\times$ 1080 & 35M    & 35590ms & 0.028  \\
    Li 2022~\cite{mm22}  &$\times$ 1080& 67M & 525ms& 1.9	 \\
    Sheng 2021~\cite{DBLP:journals/corr/abs-2111-13850}  &$\times$ 1080 &41M & 470ms& 2.12 \\

     \midrule
    HNeRV~\cite{HNeRV22} &$\times$ 960 & 3.2M & 30ms & 33.3  \\
    DNeRV &$\times$ 960 & 3.5M  & 39ms & 25.6  \\
  \end{tabular}
  \caption{Complexity comparison.}
  \label{decode}
\end{table}

\noindent \textbf{Compare with state-of-the-arts methods}. We compare DNeRV with H.264~\cite{h264}, H.265~\cite{h265}, DVC~\cite{dvc}, FVC~\cite{FVC}, M-LVC~\cite{MLVC}, DCVC~\cite{DBLP:conf/nips/LiLL21} and ELF-VC~\cite{ELFVC}. Without any specific modification, DNeRV is better than traditional video codecs H.264 or H.265 in both PSNR and SSIM, and it is also competitive with the state-of-the-art deep compression methods. We will explore the potential of DNeRV in video compression in the follow-up studies. The complexity comparison of video decoding with two rapid neural compression methods~\cite{mm22, DBLP:journals/corr/abs-2111-13850} in model size, decoding time and FPS is shown in Tab~\ref{decode}.

\subsection{Video Interpolation}
\begin{figure*}[t]
  \centering
  \includegraphics[width=0.95\linewidth]{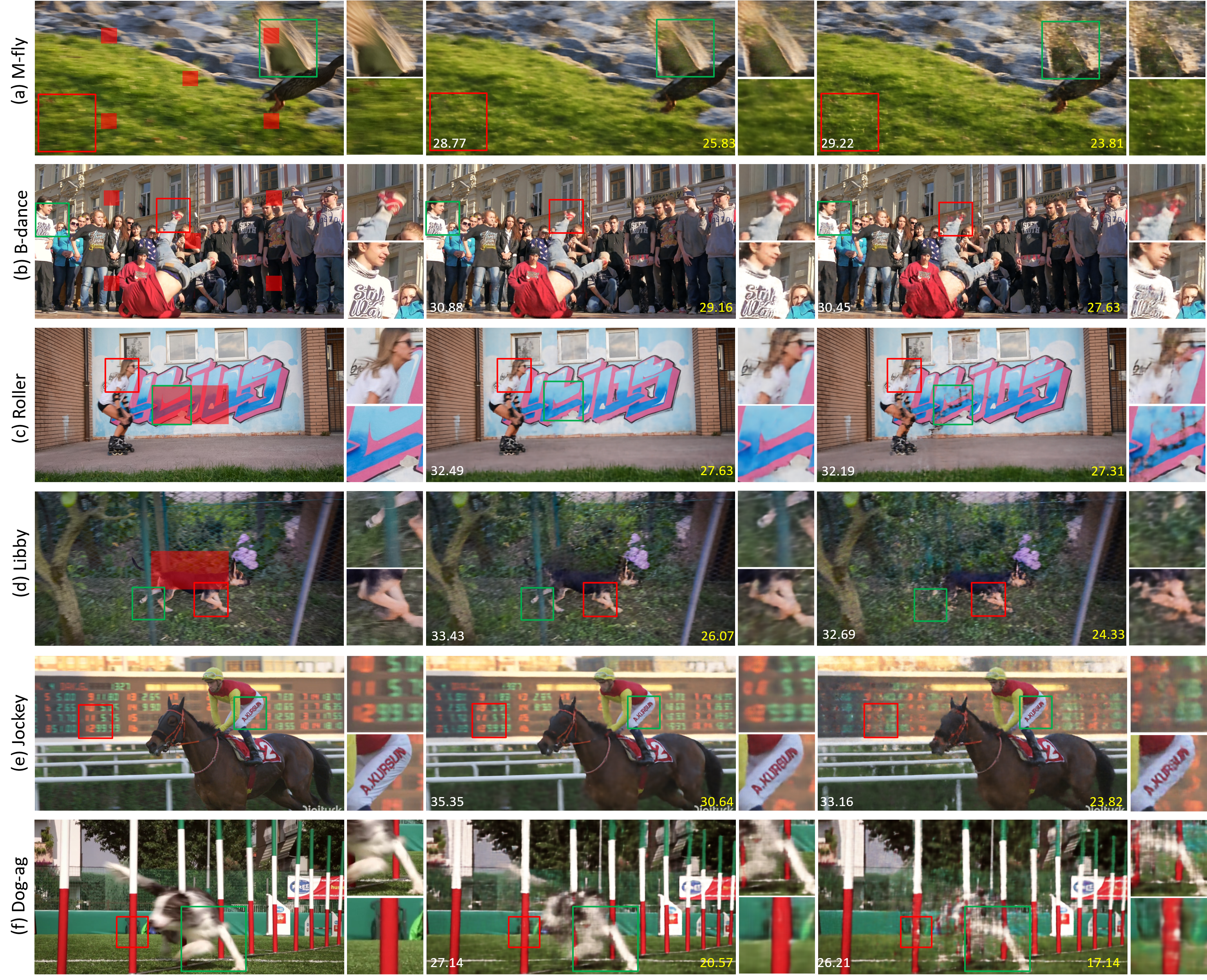}\\
  \caption{Visualization comparison for inpainting (a, b, c, d) and interpolation (e, f) results on Davis Dynamic. The left is ground truth, DNeRV's results in mid and HNeRV's on the right. White number is the best PSNR for each method training on original video, while yellow ones are testing PSNR on masked videos. Zoom in for details.} \label{inp}
\end{figure*}

Since the INR-based video representation is over-fitting to the given video, we evaluate the generalization of different methods by video interpolation task. We follow the setting in \cite{HNeRV22} and demonstrate the quantitative results on UVG in Tab.~\ref{int} and qualitative results in Fig.~\ref{inp}. Thanks to the consideration of adjacent dynamics, DNeRV outperforms other implicit methods especially on the videos owning dynamic scene and large motion. More results on DAVIS are reported in appendix.

\noindent \textbf{Case study}. Interpolation is quite challenging for the generalization ability of INRs. Shown in Tab.\ref{int}, DNeRV earns 28.5\% and 14.6\% improvement against the best results on ``Jockey" and ``ReadySetGo", where exist strong dynamics and large motion. Especially for ``Jockey" in Fig.~\ref{inp}, we could recognize some numbers or letters from DNeRV's prediction, but it's impossible for HNeRV's.
\begin{table}[t]\footnotesize
    \setlength{\tabcolsep}{1.5pt}
  \centering
  \begin{tabular}{l|lllllll|c}
     UVG         & Beauty & Bospho & Honey & Jockey & Ready & Shake &Yacht &avg.\\
    \midrule[1.5pt]
    NeRV~\cite{nerv}       &28.05  & 30.04 & 36.99          &20.00         &17.02  & 29.15  &24.50 & 26.54 \\
    E-NeRV~\cite{e-nerv}   &27.35  & 28.95 & 38.24          &19.39         &16.74  & 30.23  &22.45 & 26.19 \\
    H-NeRV~\cite{HNeRV22}  &31.10  & 34.38 & \textbf{38.83} &23.82         &20.99  & 32.61  &27.24 & 29.85 \\
 \midrule
    \textbf{D-NeRV} & \textbf{35.99} &\textbf{35.19} &37.43 &\textbf{30.61} &\textbf{24.05} &\textbf{35.34} &\textbf{28.70} &\textbf{32.47}\\
  \end{tabular}
  \caption{Video interpolation results on 960 $\times$ 1920 UVG in PSNR.}
  \label{int}
\end{table}

\subsection{Video Inpainting}
We conduct video inpainting experiments on DAVIS Dynamic with central mask and disperse mask. The central mask follows the setting in~\cite{Xu2019DeepFV, Ouyang2021InternalVI} as the rectangular mask with width and height both 1/4 of the original frame. The disperse mask is five square masks in $100 \times 100$ average distributed in fixed positions. The quantitative results are listed in Tab.~\ref{vr_davis} and qualitative performance is shown in Fig.~\ref{inp}. We train the implicit methods on raw videos but test them with masked videos. We only conduct the comparison with HNeRV because it beats other implicit methods on robust inpainting, reported in~\cite{HNeRV22}. Although the diff is also masked before input, DNeRV still obtains impressive and robust results.

\noindent \textbf{Case study}. Detailed texture is a major difficulty for DNeRV to encode videos, because difference of neighboring frames is more like than high frequency details close to noise, such as ``Mallard-fly'' and ``Cows''. However, DNeRV outperform other implicit methods in robustness and generalization. Although in ``Mallard-fly'', DNeRV's training PSNR is less than HNeRV, but severe trivial-fitting phenomenon happens. Those similar textures which in the same position among frames, seem as copies in HNeRV's results, shown as (a, d) in Fig~\ref{inp}.

\subsection{Ablation Study}
We provide various ablation studies on Bunny with PSNR and MS-SSIM as the metrics, reported in Tab~\ref{fusion} and Tab.~\ref{diff_abla}.

\noindent \textbf{Diff stream}. We compare the backward difference, forward difference, central difference and second order difference as input and merge one of them with content stream in 3rd stage. The results are shown in Tab.~\ref{diff_abla}, indicating that higher-order difference may not be helpful for fitting. We concatenate both backward and forward difference as our default diff stream.

\noindent \textbf{Param quantity}. To verify the effectiveness of diff stream, we compare the hybrid ones without diff stream in varying model size (by changing the channels in decoder) with DNeRV. More parameters may not attributed to better performance. The params, converge speed and generalization ability should be balanced.

\noindent \textbf{Diff embedding}. The smaller shape of diff embedding would reduce both the model size and reconstruction quality. The shape in $2 \times 40 \times 80$ is adopted as default.

\noindent \textbf{Fusion module}. The ablation results reported in Tab~\ref{fusion} and Tab.~\ref{diff_abla} verify the effectiveness of proposed CCU. Compared with conv fusion or sum fusion, CCU improves PSNR by adaptively merging two-stream features.

\noindent \textbf{Fusion stages}. It is vital for DNeRV to select the stage where two streams are merged, we do ablations on every possible connection. Decoder should balance the params and the difficulty of gradient descent. DNeRV adopts merging in 3rd stage without sum fusion. Further we upgrade conv fusion to CCU.

\section{Conclusion}
In this paper, we propose the difference neural representation for videos (DNeRV) for modeling the inherent dynamics of contextual frames. Relying on the diff stream and collaborative content unit, DNeRV retains its advantages in reconstruction quality for video regression, compression, inpainting and interpolation. Consequently, the experimental results
show that the proposed DNeRV could achieve effective and robust representation for videos by achieving a better approximation to the implicit mapping than existing NeRV methods.

\noindent \textbf{Future directions}. DNeRV shows its potential on various visual tasks. The improved task-specific approach based on DNeRV is promising to challenge the state-of-the-art methods. Also, rigorous theoretical analysis needs to be improved for INR-based networks $g_\theta$ fitting the continuous $f$ on finite training tuple via gradient descent.

\noindent \textbf{Acknowledgements.} We thank Xingyu Xie for comments that greatly improved the manuscript. The work was supported in part by National Key Research and Development Project of China (2022YFF0902402).

{\small
\bibliographystyle{ieee_fullname}
\bibliography{egbib}
}


{\centering\section*{Appendix}}
\section{Implementation Details}
\subsection{Architecture of diff encoder}
The strides in diff encoder depends on the resolution of input frame and the size of output diff embeddings, shown as Tab.~\ref{a_de}.
\begin{table}[h]\small
    \setlength{\tabcolsep}{5pt}
  \centering
  \begin{tabular}{ll|l}
    size of diff embeddings &resolution   & strides \\
    \midrule[1.5pt]
   2 $\times$ 40 $\times$ 80 & 640 $\times$ 1280   &(2,2,2,2)\\
    &480 $\times$ 960    &(3,2,2)  \\
    &960 $\times$ 1920   &(4,3,2)  \\
   \midrule
    2 $\times$ 10 $\times$ 20       &960 $\times$ 1920              &(4,4,3,2)\\
  \end{tabular}
  \caption{Architecture of diff encoder.}
  \label{a_de}
\end{table}

\subsection{Architecture of decoder}
The architecture of decoders with CCU in different size is given in Tab.~\ref{a_b}. $C_{init}$ is the initial channel width of embeddings before feeding to decoder stages. Once feeding the $w_0 \times h_0 \times C_{init}$ embeddings into following stage, for example Stage 1 with stride $s=5$, the size of output feature maps is $5w_0 \times 5h_0 \times C_{1}$, where $C_1 = \lfloor C_{init} /r \rfloor$ is the output channel width of Stage 1, $r=1.2$ is the reduction rate for each stage and $\lfloor x \rfloor$ is the round down operator. $K_c$ is the kernel size in CCU. The minimal and maximal kernel size in different decoder stages are 1 and 5, follow the setting in~\cite{HNeRV22}.

\begin{table}[h]\small
    \setlength{\tabcolsep}{5pt}
  \centering
  \begin{tabular}{l|ccccccc}
    resolution    &size   & $C_{0}$ & $C_{init}$ & $C_{1}$ & $C_{5}$   & strides &$K_c$ \\
    \midrule[1.5pt]
    640 $\times$ 1280 &0.35   &16   &32 &26 &11  &(5,4,4,2,2) &3\\
                   &0.75   &16   &48 &40 &18  &(5,4,4,2,2) &3\\
                   &1.5    &16   &68 &56 &25  &(5,4,4,2,2) &3\\
                   &3      &16   &95 &79 &37  &(5,4,4,2,2) &3\\
  \midrule
    480 $\times$ 960  &3     &16   &110 &91 &42  &(5,4,3,2,2) &1\\
  \midrule
    960 $\times$ 1920 &1.58   &16   &68 &56 &25  &(5,4,4,3,2) &1\\
                     &3      &16   &92 &76 &35  &(5,4,4,3,2) &1\\
  \end{tabular}
  \caption{Architecture of decoder and CCU.}
  \label{a_b}
\end{table}

\subsection{Experimental Details}
In video compression, the network structure would be adjusted for different sizes and bpp, 1.58M with diff embedding in $2 \times 10 \times 20$ for 0.0146 bpp, 3M with diff embedding in $2 \times 10 \times 20$ for 0.0257 bpp and 3M with diff embedding in $2 \times 40 \times 80$ for 0.0517 bpp.

\section{Additional quantitative results}
\subsection{Comparison for video interpolation on DAVIS Dynamic}
Interpolation results between different methods on DAVIS Dynamic are shown in Tab.~\ref{I_D}. We only compare DNeRV with hybrid-based implicit methods~\cite{HNeRV22} because HNeRV is the current best implicit method for video representation.

\subsection{The effects of different compression techniques}
Ablations for various compression technique on UVG is given in Tab.~\ref{a_c}. In future work, more advanced model compression methods would be used on the NeRV methods owing to the fewer redundance in the weights.

\subsection{The effects of different compression techniques}
For the evaluation of video compression, the results of VMAF~\cite{Rassool2017VMAFRV} are demonstrated in~Tab.~\ref{vmaf}.
\begin{table}[h]\small
    \setlength{\tabcolsep}{1.5pt}
  \centering
  \begin{tabular}{l|lllllll|c}
     Bpp         & Beauty & Bospho & Honey & Jockey & Ready & Shake &Yacht &avg.\\
    \midrule[1.5pt]
    0.015        &77.74  &71.43   &93.71  &68.02  &53.55  &80.74  &57.55  &71.82  \\
    0.025        &83.78  &78.18   &93.16  &75.38  &60.97  &82.53  &63.45  &76.78  \\
    0.05         &85.15  &77.45   &94.22  &84.02  &67.47  &86.13  &60.09  &79.22  \\
  \end{tabular}
  \caption{Number of VMAF on 960 $\times$ 1920 UVG in different Bpp.}
  \label{vmaf}
\end{table}

\subsection{Ablation results for optimizer}
Results for optimizer ablations on Bunny with 0.35M size and 300 epochs is given in Tab.~\ref{a_o}. Adan~\cite{Xie2022AdanAN} is much more effective than Adam for larger learning rate.
\begin{table*}[t]\small
    \setlength{\tabcolsep}{3pt}
  \centering
  \begin{tabular}{cc|cccccc}
    learning rate  &optimizer     &50 & 100 & 150 & 200 & 250 & 300 \\
    \midrule[1.5pt]
    5e-4       &Adam  &24.97/0.769 &27.86/0.873 &28.99/0.905 &30.10/0.920 &30.66/0.926 &30.80/0.927 \\
               &Adan  &24.17/0.734 &26.42/0.823 &27.65/0.862 &28.41/0.881 &28.80/0.890 &28.91/0.893 \\
    \midrule
    1e-3       &Adam  &26.36/0.829 &24.67/0.776 &27.06/0.841 &27.71/0.863 &28.46/0.876 &28.56/0.878 \\
               &Adan  &25.53/0.789 &28.23/0.879 &29.31/0.905 &30.03/0.917 &30.40/0.922 &30.50/0.924 \\
    \midrule
    3e-3       &Adam  &18.39/0.519 &18.81/0.548 &19.31/0.584 &19.18/0.583 &19.32/0.591 &19.36/0.594 \\
               &Adan  &27.59/0.865 &29.76/0.918 &30.59/0.933 &31.35/0.941 &31.78/0.944 &31.89/0.946 \\
  \end{tabular}
  \caption{Optimizer ablations on Bunny in PSNR/SSIM.}
  \label{a_o}
\end{table*}

\begin{table*}[t]\small
    \setlength{\tabcolsep}{3pt}
  \centering
  \begin{tabular}{l|ccccccc}
     UVG                 & Beauty & Bospho & Honey & Jockey & Ready & Shake &Yacht \\
    \midrule[1.5pt]
    N/A                           &40.00/0.972 &36.67/0.965 &41.92/0.993 &35.75/0.947 &28.68/0.917 &36.53/0.962 &31.10/0.924  \\
    8-bit Quant                   &39.97/0.972 &36.64/0.965 &41.20/0.993 &35.73/0.947 &28.66/0.916 &36.35/0.961 &31.00/0.923\\
    8-bit Quant + Pruning (10\%)  &39.38/0.971 &36.41/0.964 &39.95/0.991 &35.50/0.946 &28.55/0.915 &35.42/0.959 &30.78/0.921 \\
    8-bit Quant + Pruning (20\%)  &33.72/0.961 &34.56/0.957 &34.47/0.978 &32.25/0.938 &27.63/0.905 &28.66/0.943 &28.84/0.908\\
  \end{tabular}
  \caption{Compression ablations on UVG in PSNR/SSIM.}
  \label{a_c}
\end{table*}

\begin{table*}[t]\small
  \centering
  \begin{tabular}{l|cc|cc}
   \multirow{2}*{\shortstack{Videos }} &\multicolumn{2}{c|}{DNeRV} &\multicolumn{2}{c}{HNeRV} \\
     &test & train  &test & train\\
     \midrule[1.5pt]
    Blackswan &23.89/0.712 &28.98/0.874  &21.67/0.589 &28.76/0.865  \\
    Bmx-bumps &22.34/0.696 &25.96/0.784  &19.24/0.549 &30.32/0.883  \\		
    Camel     &21.31/0.656 &23.79/0.761  &20.69/0.586 &26.28/0.855  \\			
    Breakdance&22.28/0.858 &27.26/0.937  &20.40/0.841 &29.53/0.958	\\	
    Car-round &20.42/0.725 &28.91/0.931  &16.92/0.560 &28.23/0.919  \\
    Bmx-trees &21.68/0.644 &28.88/0.867  &18.39/0.453 &28.99/0.872  \\		
    Car-shadow&22.47/0.734 &29.41/0.913  &19.35/0.622 &28.64/0.897  \\
    Cows      &20.89/0.629 &25.24/0.837  &20.45/0.590 &24.71/0.815	\\
    Dance-twirl&20.95/0.656&29.19/0.872  &18.38/0.517 &28.70/0.857  \\
    Dog       &24.91/0.683 &29.55/0.857  &21.99/0.457 &29.85/0.868  \\
    Car-turn  &24.29/0.737 &28.21/0.838  &22.34/0.654 &27.80/0.828  \\
    Dog-agility&20.57/0.730&27.14/0.852  &17.14/0.609 &26.21/0.818  \\
    Drift-straight&19.11/0.645&29.75/0.921&15.62/0.354&29.72/0.916  \\
    Drift-turn&21.22/0.649 &29.45/0.849  &18.44/0.501 &28.43/0.815  \\
    Goat      &20.46/0.554 &28.63/0.908  &18.22/0.327 &27.69/0.891  \\
    Libby     &24.24/0.688 &32.22/0.906  &20.00/0.472 &30.75/0.871  \\
    Mallard-fly&21.81/0.610&28.25/0.809	 &19.23/0.397 &27.26/0.788  \\
    Mallard-water &21.24/0.687 &27.55/0.882	&17.60/0.429 &29.23 0.911\\
    Parkour   &22.13/0.680 &27.32/0.879  &18.82/0.488 &26.77/0.863  \\
    Rollerblade&24.91/0.850&30.52/0.915  &21.56/0.782 &29.92/0.907  \\	
    Scooter-black&17.15/0.633 &27.26/0.926&14.37/0.416&26.33/0.901 \\	
    Stroller   &23.32/0.718 &32.36/0.923 &20.47/0.559 &31.68/0.905 \\
  \midrule
   Average     &\textbf{21.89/0.690} &28.45/0.875 &19.15/0.534 &28.44/0.873 \\
  \end{tabular}
  \caption{Interpolation results on DAVIS Dynamic.}
  \label{I_D}
\end{table*}
\begin{figure*}[h]
  \centering
  \includegraphics[width=\linewidth]{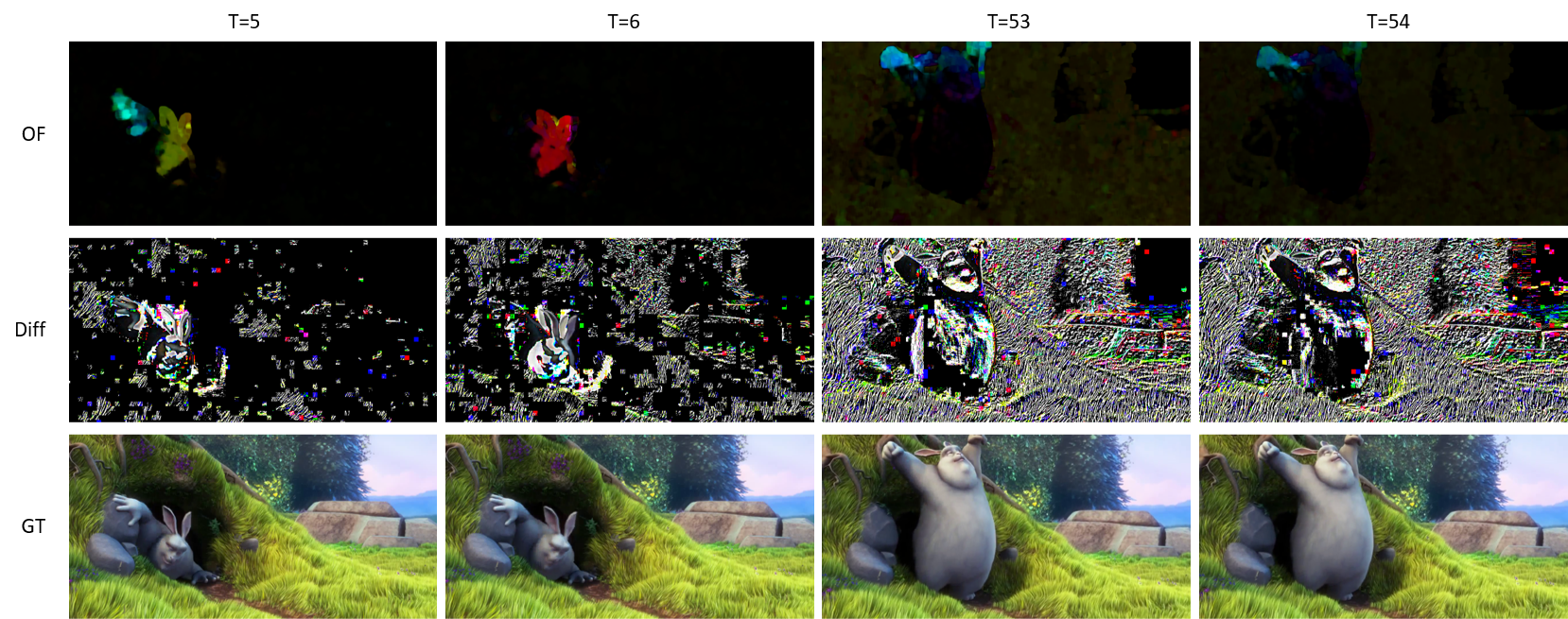}\\
  \caption{Comparison between optical flow and difference stream.}
  \label{of}
\end{figure*}

\section{Additional qualitative results}
\subsection{Visualization of video interpolation on UVG}
\begin{figure*}[t]
  \centering
  \includegraphics[width=\linewidth]{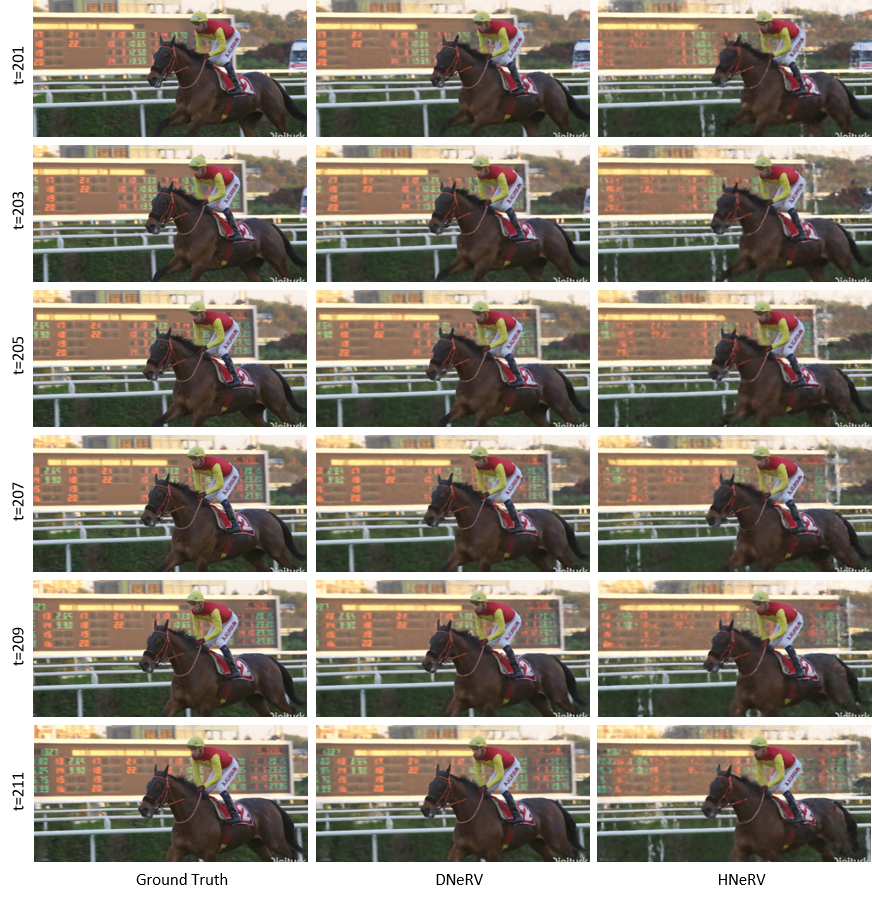}\\
  \caption{Additional examples for video interpolation on Jockey.}
  \label{int_u_1}
\end{figure*}

\begin{figure*}[t]
  \centering
  \includegraphics[width=\linewidth]{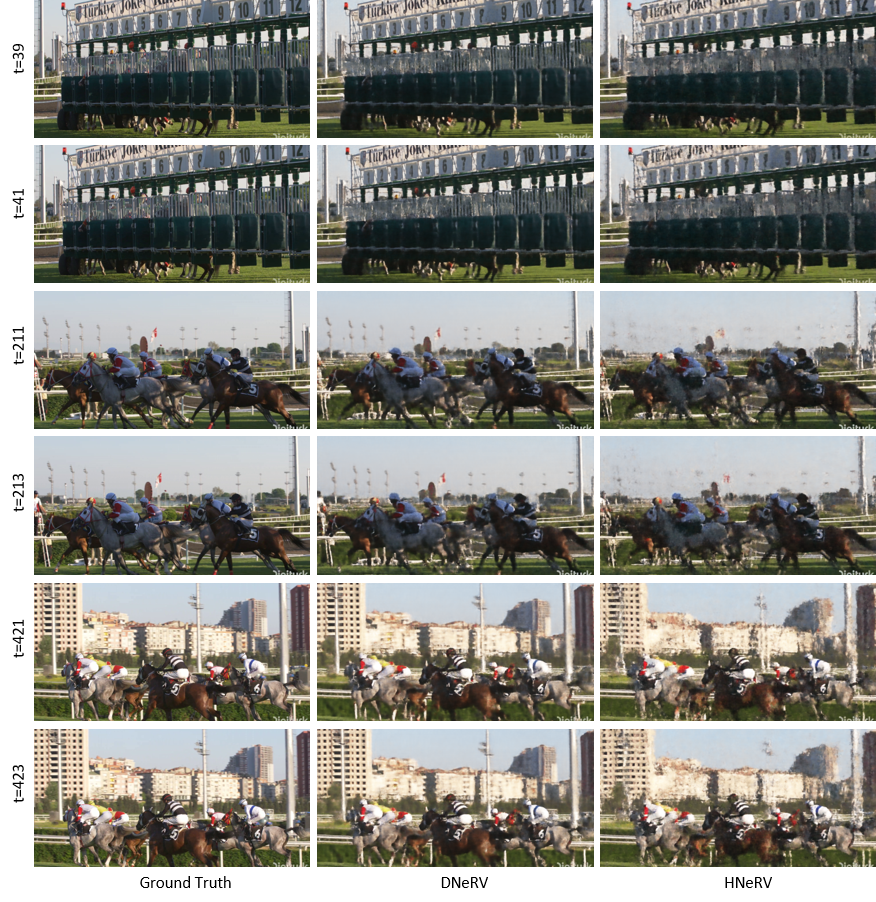}\\
  \caption{Additional examples for video interpolation on ReadySetGo.}
  \label{int_u_2}
\end{figure*}

Additional interpolation comparison on UVG is given in Fig.~\ref{int_u_1} and Fig.~\ref{int_u_2}.

``Jockey'' and ``ReadySetGo'' are two typical videos with large motion and dynamic scenes from UVG. In Fig.~\ref{int_u_1} and Fig.~\ref{int_u_2}, we could find that the interpolations generated by DNeRV are obviously better than HNeRV. Some subtle spatial structures in interpolations of DNeRV, such as numbers on the screen or flagpole in the distance, remain nearly constant between adjacent frames.

\subsection{Visualization of video interpolation on DAVIS Dynamic}
\begin{figure*}[t]
  \centering
  \includegraphics[width=\linewidth]{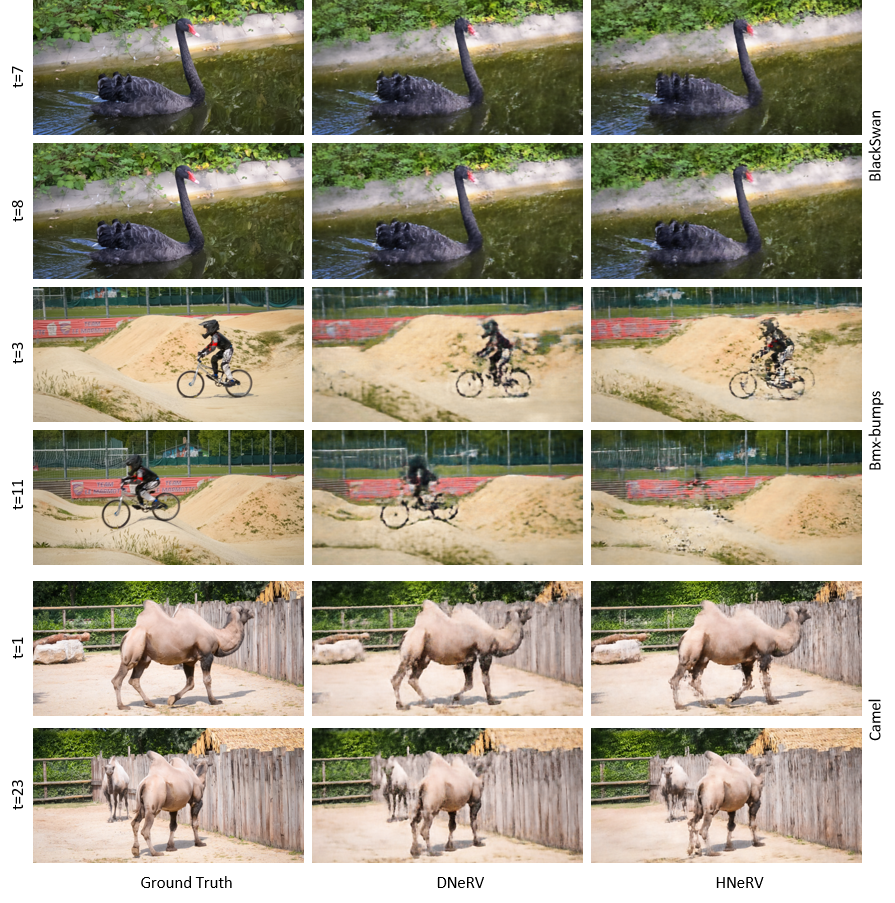}\\
  \caption{Additional examples for video interpolation on Blackswan, Bmx-bumps and Camel.}
  \label{int1}
\end{figure*}

\begin{figure*}[t]
  \centering
  \includegraphics[width=\linewidth]{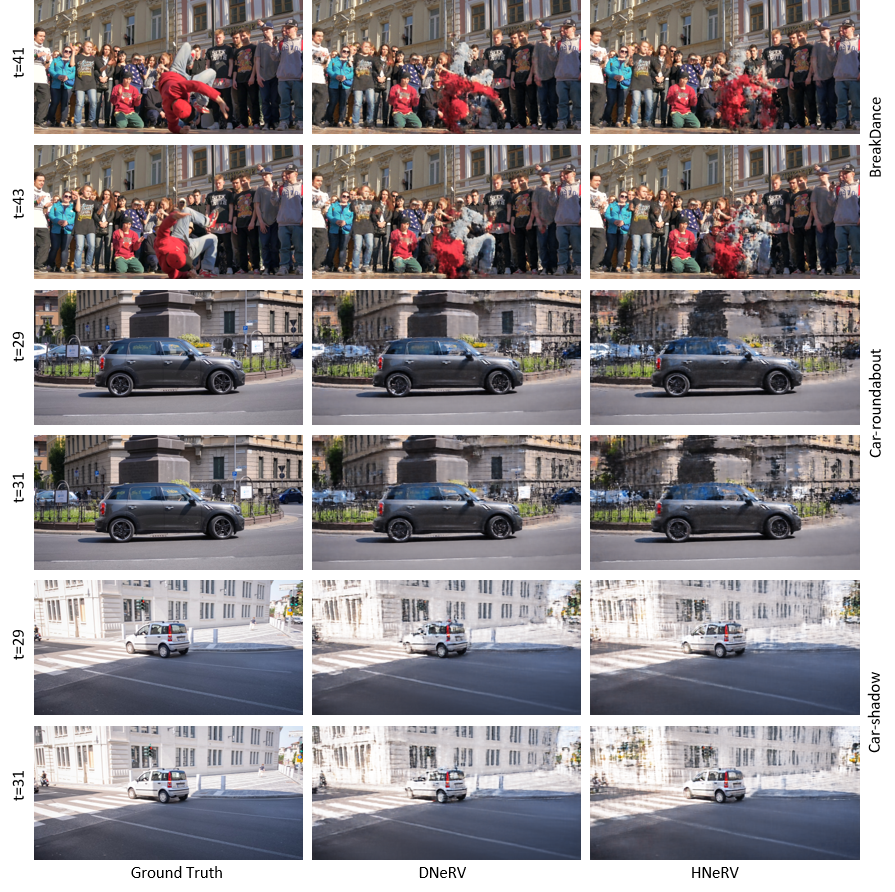}\\
  \caption{Additional examples for video interpolation on Breakdance, Car-roundabout and Car-shadow.}
  \label{int2}
\end{figure*}

\begin{figure*}[t]
  \centering
  \includegraphics[width=\linewidth]{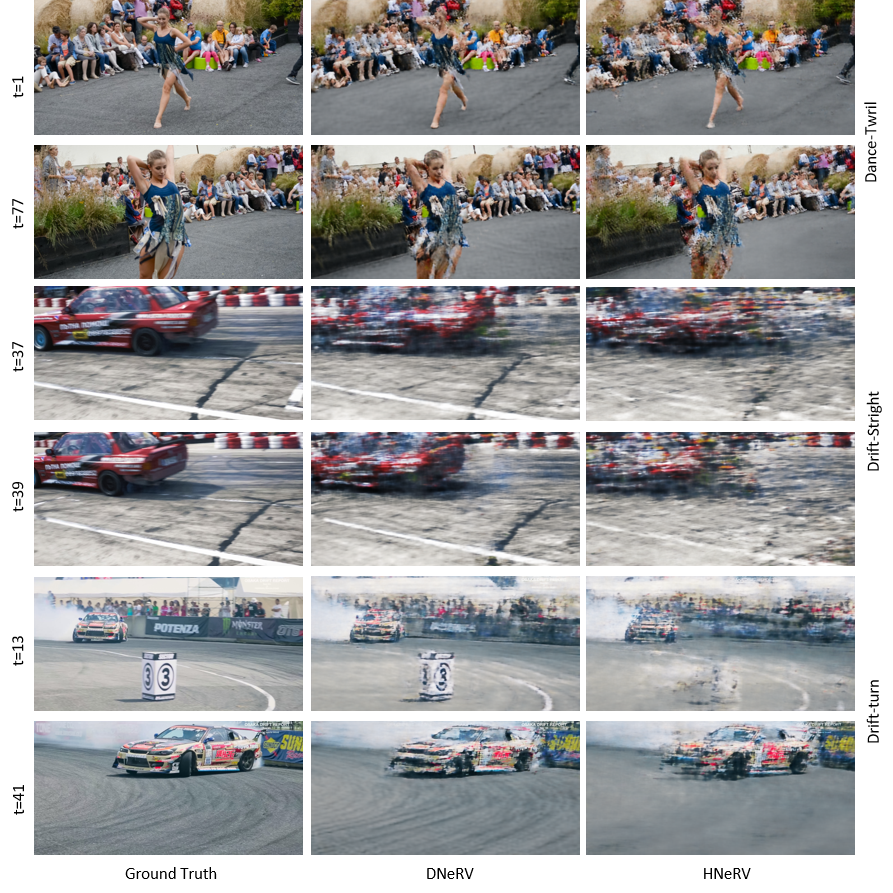}\\
  \caption{Additional examples for video interpolation on Dance-twril, Drift-straight and Drift-turn.}
  \label{int3}
\end{figure*}

\begin{figure*}[t]
  \centering
  \includegraphics[width=\linewidth]{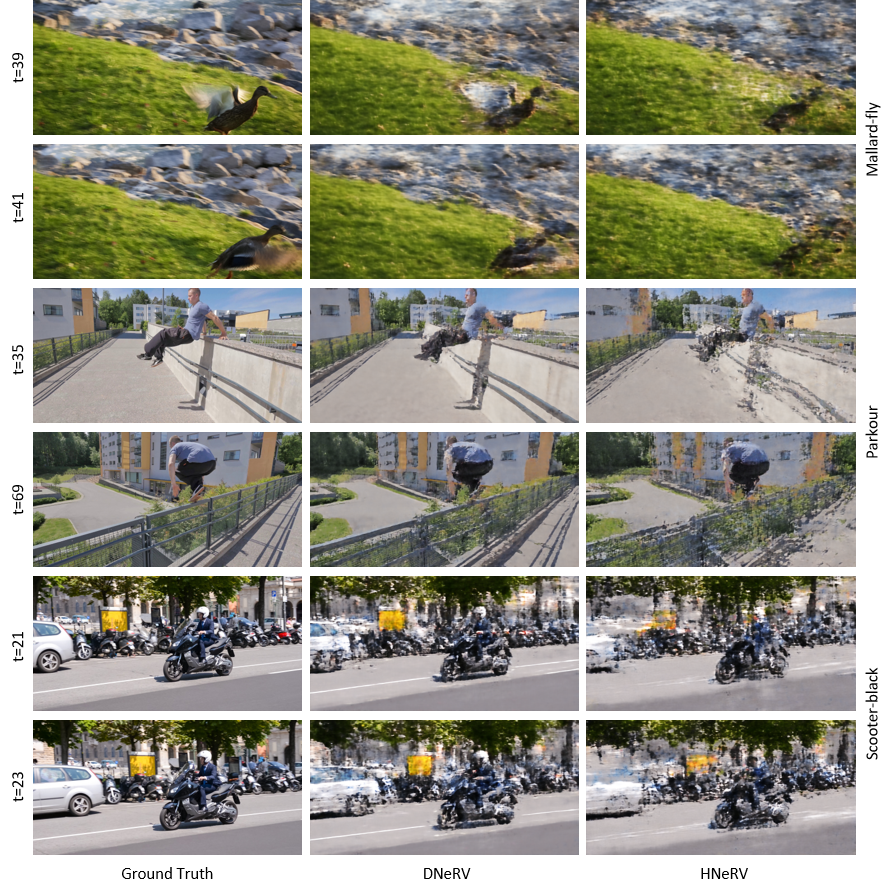}\\
  \caption{Additional examples for video interpolation on Mallard-fly, Parkour and Scooter-black.}
  \label{int4}
\end{figure*}

Additional interpolation comparison on DAVIS Dynamic is given in Fig.~\ref{int1}, Fig.~\ref{int2}, Fig.~\ref{int3} and Fig.~\ref{int4}.

DAVIS Dynamic is more difficult than UVG by reason of more dynamic scene changing and fewer frames. Although DNeRV outperforms HNeRV achieving the best results of implicit methods, but there is still much room for improvement. Once increasing the parameter quantity and utilizing task-specific modification, DNeRV could be competitive with state-of-the-art deep interpolation methods.

\subsection{Visualization of video inpainting on DAVIS Dynamic}
\begin{figure*}[t]
  \centering
  \includegraphics[width=\linewidth]{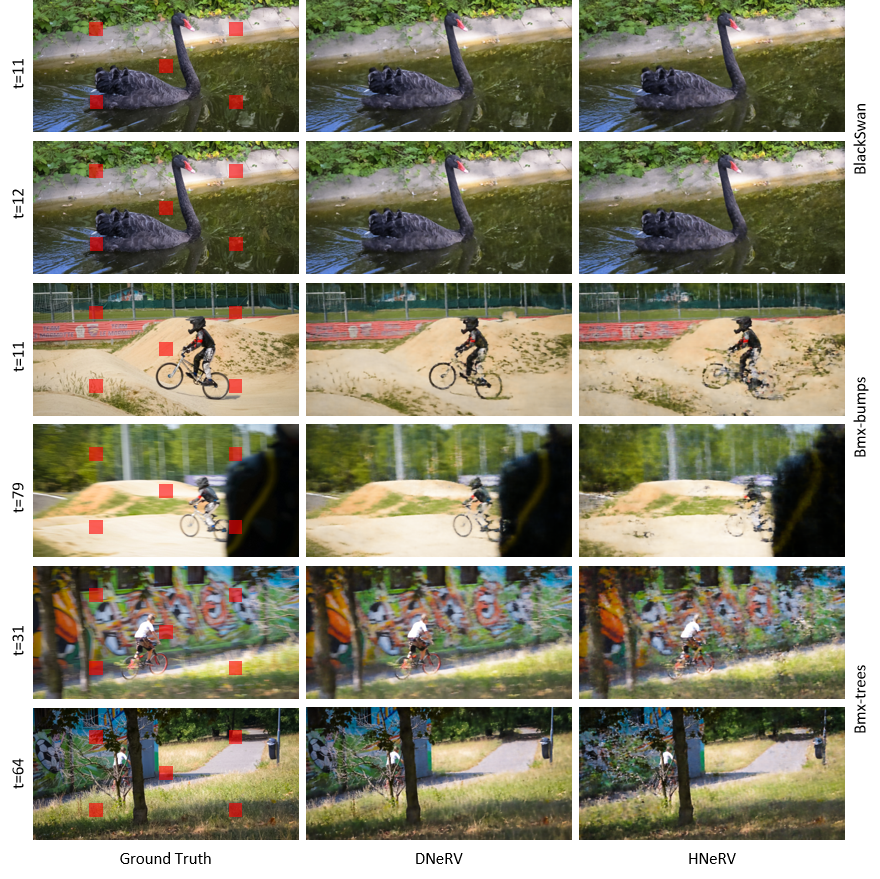}\\
  \caption{Additional examples for video interpolation on Blackswan, Bmx-bumps and Bmx-trees.}
  \label{inp1}
\end{figure*}

\begin{figure*}[t]
  \centering
  \includegraphics[width=\linewidth]{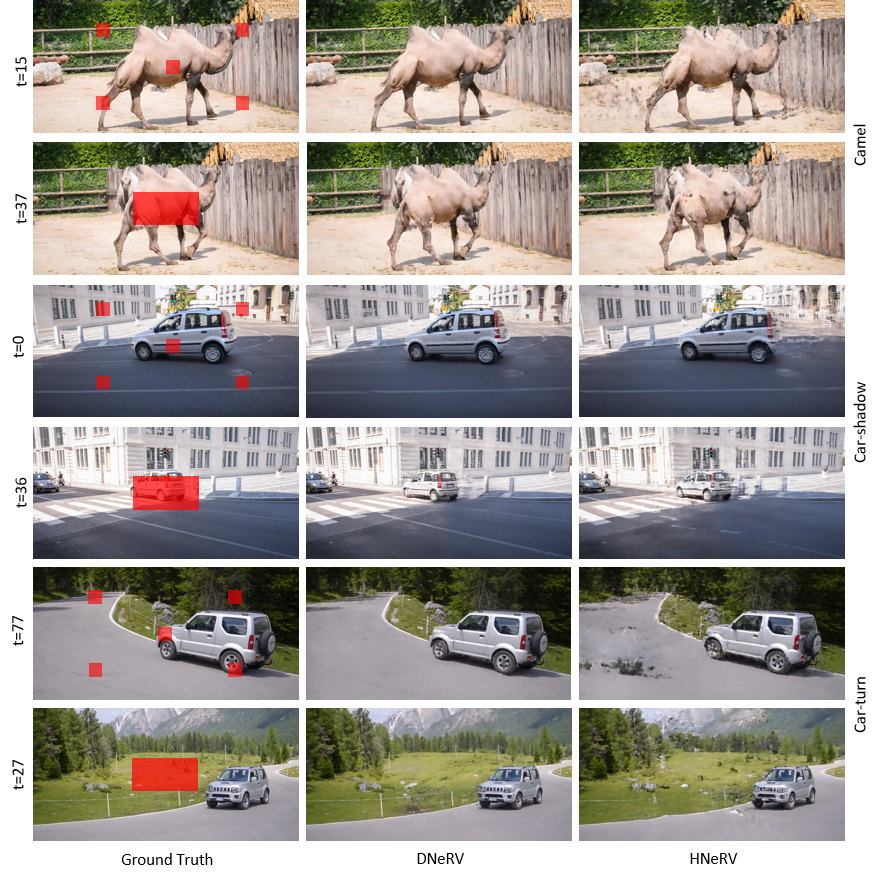}\\
  \caption{Additional examples for video interpolation on Camel, Car-shadow and Car-turn.}
  \label{inp2}
\end{figure*}

\begin{figure*}[t]
  \centering
  \includegraphics[width=\linewidth]{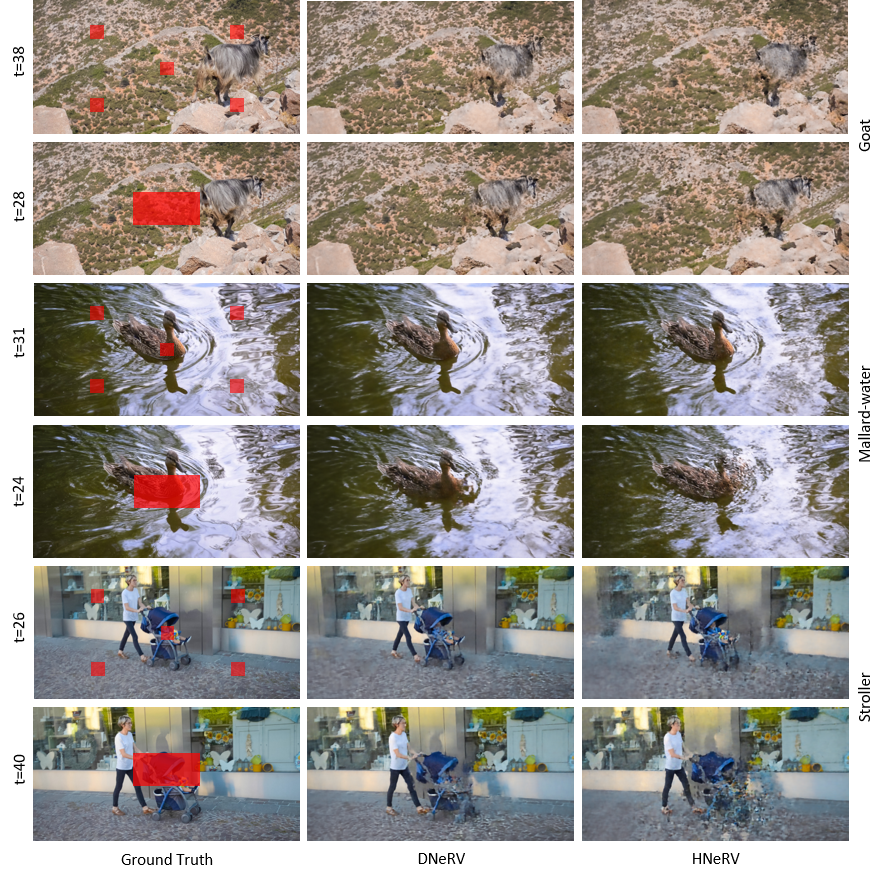}\\
  \caption{Additional examples for video interpolation on Goat, Mallard-water and Stroller.}
  \label{inp3}
\end{figure*}

Additional inpainting comparison on DAVIS Dynamic is shown in Fig.~\ref{inp1}, Fig.~\ref{inp2} and Fig.~\ref{inp3}.

Due to diff stream and CCU, DNeRV could model different regions of the frame more robustly, reduce the influence of masked regions. Besides, one limitation of DNeRV is that it couldn't model the detail texture well, and we will improve it in the future work.

\subsection{Visualization of optical flow and difference stream}
We conducted additional experiments on Bunny, following the same setting as Tab.~\ref{vr_size}. The PSNR results are 29.13, 29.25, 28.84, and 28.70 in dB for the model sizes of 0.35M, 0.75M, 1.5M, and 3M. The optical flow is computed using Gunner Farneback algorithm by opencv-python 4.5.3 and numpy 1.19.5.

The visualization comparison between optical flow and diff stream is shown in Fig.~\ref{of}. It can be clearly observed that, although optical flow contains motion information, it loses huge other information in pixel domain. Saliency motion information in optical flow may be key in action recognition or motion prediction, but it cannot bring much help for pixel-level reconstruction tasks. For example, the fluctuation of grass or the change of skin brightness with the light may not help to recognize the rabbit's movements, but they are essential for reconstruction. Diff stream records all these information in unbiased way.

\end{document}